%% file: main.tex
\documentclass[10pt, logo]{hyreport}

\usepackage{multirow}
\usepackage{xspace}
\usepackage{enumitem}
\usepackage{graphicx}
\usepackage{amsmath} 
\usepackage{multirow}
\usepackage{colortbl}
\usepackage{adjustbox} 
\usepackage{subcaption}
\usepackage{pifont}
\usepackage{wrapfig}
\usepackage{amssymb} 
\usepackage{booktabs} 
\usepackage{algorithm}
\usepackage{makecell}
\usepackage[table]{xcolor}
\usepackage{tabularx}
\usepackage{booktabs}
\usepackage{threeparttable}
\usepackage{diagbox}
\usepackage[most]{tcolorbox}
\usepackage{booktabs,subcaption}
\definecolor{darkred}{HTML}{9B1C31}

\definecolor{googleRed}{HTML}{EA4335}
\definecolor{googleBlue}{HTML}{4285F4}
\definecolor{googleGreen}{HTML}{34A853}
\definecolor{googleYellow}{HTML}{FBBC05}

\newsavebox{\panelbox}

\newtcolorbox{findingbox}[1][Key finding]{
  enhanced, breakable, arc=4pt, boxrule=0.6pt,
  colback=black!3, colframe=themecolor!70,
  left=8pt, right=8pt, top=4pt, bottom=4pt,
  before skip=8pt, after skip=8pt,
  before upper={\textbf{\textcolor{themecolor}{#1.}}\enspace}
}

\renewcommand{\appendixtitle}{%
  \clearpage
  {\centering\titlefont Appendix\par}%
  \vskip 8pt}

\title{One Prior, Two Dynamics: Repurposing Video Diffusion\\
Transformers for Self-Supervised Latent Action Learning}
\shorttitle{Repurposing Video Diffusion Transformers for Self-Supervised Latent Action Learning}

\author{
  {\small Yuxin Jiang$^{1,2}$, Wenqiang Sun$^{3}$, Jiaqi Li$^{3}$, Yuchao Gu$^{1}$, Lvmin Zhang$^{4}$} \\
  {\small \textbf{Ivor Tsang$^{2}$, Mike Zheng Shou$^{1}$}} \\~\\
  $^{1}$Show Lab, National University of Singapore \quad $^{2}$A*STAR, Singapore \\
  $^{3}$Tencent \quad $^{4}$Stanford University \\
  \url{https://showlab.github.io/o2laf}
}

\input{fig/teaser}      
\input{sec/0_abstract}

\begin{document}
\maketitle

\input{sec/1_intro}
\input{sec/2_related}
\input{sec/3_method}
\input{sec/4_exp}

\input{sec/5_analysis}

\input{sec/6_conl}

\newpage

\bibliography{ref}

\newpage
\appendix
\input{sec/7_supp}

\end{document}

%% file: fig/teaser.tex
\teaser{%
  \centering
  \includegraphics[width=\linewidth]{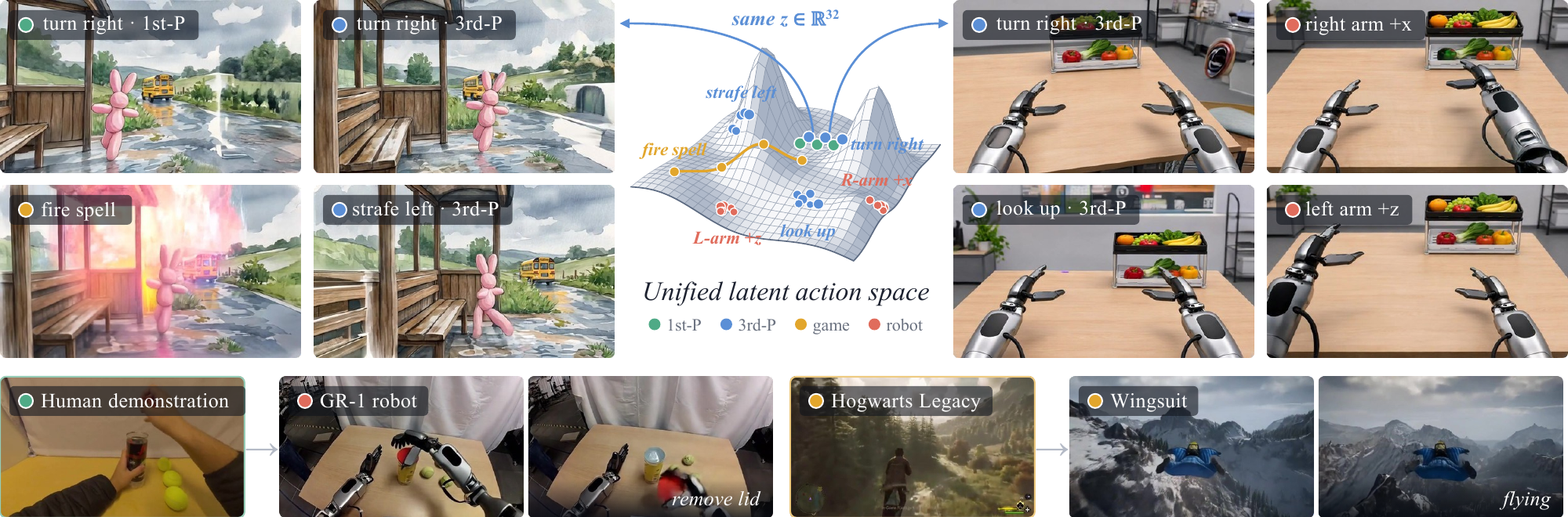}
\captionof{figure}{
\textbf{O$^2$LaF:} \underline{One} Prior, \underline{Two} Dynamics --- a self-supervised
\underline{L}atent \underline{A}ction learning \underline{F}ramework.
Latent actions from unlabeled videos share one unified space (center): e.g., the same $z$ inferred
from 1st- and 3rd-person videos drives \emph{turn right} in both navigation and robot scenes (top).
They also transfer zero-shot across embodiments and domains  (bottom).
}
  \label{fig:teaser}
  \vspace{-5mm}
}

%% file: sec/0_abstract.tex
\begin{abstract}
Videos reveal how the world changes, but rarely the actions that cause those changes.
Latent action learning recovers such supervision from visual transitions, yet often learns the
underlying dynamics largely from scratch.
We ask whether a pretrained video generator can
instead provide the prior for both inverse and
forward dynamics.
We introduce \textbf{O$^{2}$LaF}, built on a simple principle:
\emph{one prior, two dynamics}.
We initialize both models from the same pretrained video DiT and adapt them to their distinct roles:
a feed-forward transition encoder with learnable
action queries for inverse dynamics, and a causal,
action-conditioned video generator for forward dynamics.
By controlling information access and optimizing both dynamics end-to-end  through a compact bottleneck, O$^{2}$LaF learns transferable actions from unlabeled videos.
Across navigation and robotic manipulation, O$^{2}$LaF sets a new state of the art in few-shot action probing and enables stronger zero-shot action transfer across visual contexts and embodiments. 
Beyond generative video priors, we further extend
our framework to predictive representation priors
such as V-JEPA2, achieving competitive performance.
Our findings suggest a broader paradigm for latent-action learning: repurposing pretrained video models to discover transferable actions and predict their consequences.
\end{abstract}

%% file: sec/1_intro.tex



\section{Introduction}
\label{intro}

Videos provide abundant observations of how the world changes, but rarely
reveal the actions that cause those changes~\cite{baker2022vpt, schmidt2024lapo, bruce2024genie}.
This missing supervision limits our ability to learn controllable world models~\cite{ha2018worldmodels, hafner2025dreamerv3, yang2024unisim, wu2024ivideogpt, parkerholder2025genie3, nvidia2025cosmos, robbyant2026lingbotworld, chen2026hy}
and embodied agents~\cite{zitkovich2023rt2, sima2025sima2, black2025pi0, nvidia2025gr00tn1, gemini2025robotics} from internet-scale video.

Latent action learning~\cite{edwards2019ilpo, rybkin2019learning, schmidt2024lapo} offers a way to recover such supervision directly from
visual transitions.
A latent action model typically consists of an inverse dynamics model (IDM),
which infers a latent action from an observed transition, and a forward dynamics
model (FDM), which predicts its consequence:
\[
    z_t = \mathrm{IDM}(x_t,x_{t+1}),
    \qquad
    \hat{x}_{t+1} = \mathrm{FDM}(x_t,z_t).
\]
A compact bottleneck limits what $z_t$ can encode, encouraging it to retain
 important information to explain visual changes.
Learning useful latent actions therefore depends on both dynamics:
the IDM must infer the transition, while the FDM must depend on the inferred
action to predict its consequence.

\input{fig/paradigm}

Existing methods differ in where pretrained priors enter this pipeline
(Fig.~\ref{fig:paradigm}), leaving a gap on each side.
On the \emph{inverse} side, IDMs are learned from scratch, either on raw
pixels~\cite{ye2024lapa, bruce2024genie, gaoadaworld} or on top of frozen pretrained
features~\cite{bu2025univla, garrido2026learning, nie2026lary}.
They must learn to infer actions from transitions largely through the FDM's
reconstruction supervision, and such specialized latent action models are 
outperformed by general visual foundation models~\cite{nie2026lary}.
Olaf-World~\cite{jiang2026olafworld} aligns latent actions with an external video
representation to improve transferability, but its IDM is still learned from scratch.
On the \emph{forward} side, recent methods adopt a pretrained video generator as the
FDM~\cite{ren2026videoworld2, wang2026coevolving}, whose initial-frame
conditioning supplies appearance, reducing its leakage into the latent actions.
However, they keep the standard diffusion recipe, under which the generator can learn
to disregard the latent action~\cite{wang2026coevolving}, underusing its prior as a learning signal.
Yet modern video Diffusion Transformers (video DiTs)~\cite{peebles2023dit} are pretrained at scale to model how visual states evolve over time~\cite{yang2025cogvideox, polyak2024moviegen, kong2024hunyuanvideo, wan2025wan, seedance2026seedance2, brooks2024sora, minimax2026h3}.
Their representations encode rich temporal structure, while their generative
dynamics learn to predict complex visual changes.
Much of the prior needed for latent action learning therefore already
exists inside a pretrained video generator.
This raises a simple question:
\emph{can a pretrained video DiT serve both inverse and forward dynamics for
latent action learning?}
Doing so is nontrivial: the two dynamics place different demands on the same video
prior, requiring careful control over what each can access.

We introduce \textbf{O$^{2}$LaF}, a self-supervised latent-action framework
built on a simple principle: \emph{one prior, two dynamics}.
We initialize the IDM and FDM from the same pretrained video DiT and adapt them independently for their respective roles.
For \textbf{inverse dynamics}, we repurpose the video DiT as a feed-forward transition encoder, where action queries extract one compact latent action per visual transition under causal temporal attention.
For \textbf{forward dynamics}, we retain the generative role of the video DiT and introduce causal joint prediction, which predicts the future state sequence jointly, conditioned on the inferred latent actions, while preserving temporal causality.
Unlike visual synthesis, where training often emphasizes intermediate-noise timesteps~\cite{esser2024scaling, diffsynthstudio}, the FDM here also provides the learning signal that shapes the latent-action space.
At these partially noisy timesteps, the noisy future still reveals much of the target, allowing the FDM to predict the future with little reliance on the latent action.
We therefore train at high noise, which suppresses this target leakage and forces the FDM to depend on the inferred action.
The two dynamics are optimized end-to-end through a compact continuous latent-action space.

We further systematically study how the two dynamics shape the learned action space.
For example, causal joint prediction outperforms autoregressive prediction by removing clean-history shortcuts, while role-specific noise levels reveal distinct information requirements for the two dynamics.
Beyond generative video priors, we also extend the repurposing principle to self-supervised predictive representation models~\cite{assran2025vjepa2} and language-supervised video multimodal models~\cite{li2026videochat3}
as alternative IDM backbones.
Interestingly, stronger action probing performance
does not necessarily translate into better downstream
generative control, highlighting the importance of
validating latent actions beyond probing through
downstream applications.

We evaluate O$^{2}$LaF across camera navigation and robotic manipulation.
It sets a new state of the art in few-shot action probing across both discrete and continuous action spaces, indicating stronger alignment with ground-truth actions.
Beyond probing, the learned actions also transfer without target adaptation, enabling zero-shot control across first- and third-person navigation  and human-to-robot transfer in manipulation.
The causal FDM readily serves as a video world model, while its learned action-conditioned dynamics provide a strong initialization for higher-fidelity two-stage deployment.
Our contributions are threefold:
\begin{itemize}

\item
We introduce \textbf{O$^{2}$LaF}, a self-supervised latent-action framework
that repurposes a pretrained video DiT into end-to-end trained inverse and
forward dynamics models.

\item
Through a systematic study, we show that latent-action learning depends
critically on how the two dynamics are factorized and what information they receive.

\item
O$^{2}$LaF achieves state-of-the-art few-shot action probing across camera navigation and robotic manipulation, while demonstrating strong zero-shot action transfer across visual contexts and embodiments and reusable action-conditioned dynamics for world modeling.

\end{itemize}

%% file: fig/paradigm.tex
\begin{figure*}[t]
\includegraphics[width=\linewidth]{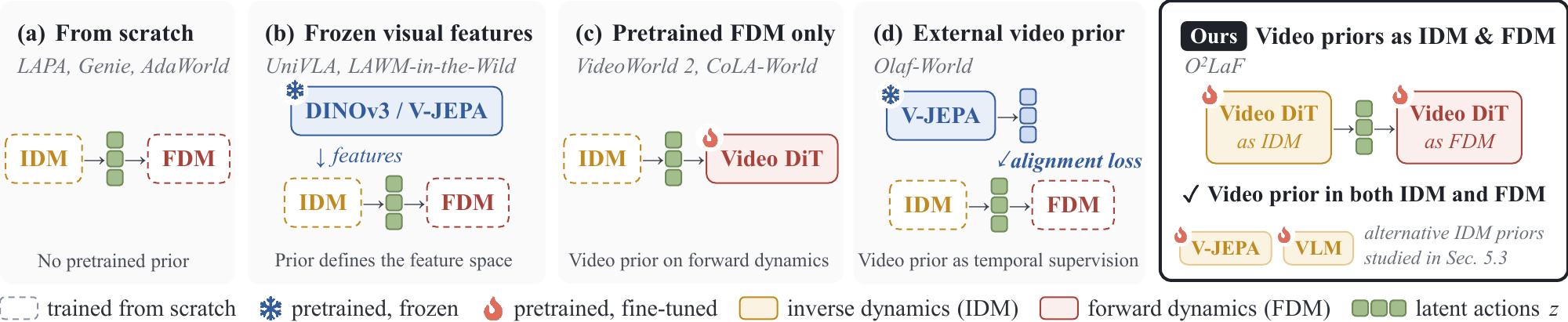}
\caption{
\textbf{Paradigms for self-supervised latent-action learning.}
Existing methods learn latent actions
(a) from scratch~\cite{bruce2024genie,ye2024lapa,gaoadaworld},
(b) on frozen visual features~\cite{bu2025univla,garrido2026learning},
(c) with a pretrained video generator as the FDM only~\cite{ren2026videoworld2,wang2026coevolving},
or (d) with an external video prior as auxiliary supervision~\cite{jiang2026olafworld}.
\textbf{Ours} repurposes one pretrained video DiT as both the IDM and the FDM.
}
\vspace{-3mm}
\label{fig:paradigm}
\end{figure*}

%% file: sec/2_related.tex
\section{Related Work}

\subsection{Latent Action Learning from Videos}

Latent action learning turns large-scale action-unlabeled video into a reusable intermediate control representation.
The same representation can be consumed by a world model to control future visual evolution~\cite{bruce2024genie, gaoadaworld, jiang2026olafworld, garrido2026learning, gao2026dreamdojo}, or produced by a policy or planner as a surrogate for physical actions~\cite{schmidt2024lapo, ye2024lapa, bu2025univla, chen2026villax, chen2024moto, yao2026unilap}.
Recent work further seeks to make this control space transferable across contexts and embodiments~\cite{jiang2026olafworld, huang2026lacwm, bauer2026lad, kim2025uniskill}.

Existing methods differ in \emph{where pretrained priors enter the
inverse--forward dynamics pipeline} (Fig.~\ref{fig:paradigm}).
Early approaches learn both transition inference and prediction directly from the target videos~\cite{schmidt2024lapo,bruce2024genie,ye2024lapa,gaoadaworld}.
To reduce shortcuts from raw-pixel reconstruction, later methods move the prediction target into pretrained semantic feature spaces, emphasizing motion and task-relevant change over appearance details
~\cite{chen2024moto,bu2025univla,wang2026repwam}.
A related direction strengthens the forward model with pretrained video priors, reducing appearance burden or replacing the from-scratch FDM~\cite{ren2026videoworld2,wang2026coevolving, liu2026scar}.
Pretrained video representations have also been used as external temporal supervision to orient latent actions across visual contexts
~\cite{jiang2026olafworld}.

Orthogonal to where the prior enters, prior work also constrains the latent
space through discrete or continuous bottlenecks
~\cite{garrido2026learning, liu2026scar, bu2026whatmatters},
semantic or cross-view objectives~\cite{lee2026mvplam, zhang2026clap, klepach2026objectcentric, jeong2026viewinvariant},
motion and consistency supervision~\cite{bi2026motus, bu2026laof, yang2026como, dai2026conla},
and physical or geometric grounding~\cite{chen2026villax, yang2026lawm3d}.
These constraints are complementary to the structural question of how the dynamics themselves are initialized.
Motivated by the shared need for transition reasoning in both inverse and forward dynamics, O$^{2}$LaF repurposes the same pretrained video DiT for both.


\subsection{Repurposing Pretrained Video Diffusion Transformers}

Pretrained video generative models encode reusable spatiotemporal structure
beyond synthesis.
Prior work exploits their internal representations for correspondence,
tracking, and geometry
~\cite{difftrack,ditflow,ditracker}, while more recent methods repurpose
the generative backbone itself for feed-forward perception and temporal
reasoning~\cite{wang2026genception,tinyhistory,nam2026trackcraft3r}.
Together, these works show that video generative pretraining provides reusable
temporal computation, not only a synthesis model.

This repurposing is increasingly extending to action learning.
Recent robot-policy and world-action models use pretrained video generators as
predictive representations, policy backbones, or priors for action experts
~\cite{vpp,cosmospolicy,dreamzero,ma2026dit4dit, fastwam, vpp2}.
These methods transfer video priors into learning a physical action space
grounded by robot supervision.
O$^{2}$LaF instead uses the video prior to discover the action representation
itself from action-free video.

%% file: sec/3_method.tex
\section{Method}
\label{method}








We introduce \textbf{O$^{2}$LaF}, a self-supervised latent action learning framework.
Our key premise is that large-scale video pretraining already captures how visual
states evolve, which both inverse and forward dynamics need.
We therefore initialize both from the same pretrained prior, but adapt each to its
role by controlling what information it can access.

\subsection{Preliminaries and Motivation}
\label{sec:prelim}

\noindent\textbf{Pretrained Video DiT.}
O$^{2}$LaF builds on a pretrained text-to-video model~\cite{wan2025wan, wan2025wan22} consisting of a video VAE,
a text encoder, and a video DiT.
The VAE maps an RGB video $\mathbf V$ to $K{+}1$ spatiotemporal latent states
$\mathbf x=(x_0,\ldots,x_K)$.
Under rectified flow~\cite{lipman2023flowmatching,esser2024scaling},
a clean video latent $\mathbf x$ is interpolated with
Gaussian noise $\boldsymbol\epsilon\sim\mathcal N(0,I)$ as
$\mathbf x_t=(1-t)\mathbf x+t\boldsymbol\epsilon$,
with target velocity $\mathbf v^\star=\boldsymbol\epsilon-\mathbf x$.
Here, $t\in[0,1]$, where $t=0$ corresponds to clean video
latents and $t=1$ to pure Gaussian noise.

\input{fig/motivation}

\noindent\textbf{Action-aware video prior.}
Prior work~\cite{difftrack, nam2026trackcraft3r, ditflow, ditracker} has shown that pretrained video DiTs contain strong temporal correspondence, enabling subjects or points to be tracked across video frames.
We ask a different question: \emph{do these models also attend to regions relevant to the ongoing action?}
We conduct a simple diagnostic. We feed clean video latents to a frozen video DiT at $t=0$ with the text
condition ``What is the action in the video?'' and
visualize the resulting cross-attention.
As shown in Fig.~\ref{fig:motivation}, the responses follow the active hands
and manipulated regions, shifting as the interaction changes across frames and clips.
Although qualitative, this observation motivates repurposing the pretrained video prior for transition-level action inference.

\subsection{Problem Formulation: One Prior, Two Dynamics}
\label{sec:formulation}

Given an action-unlabeled video, our goal is to infer one compact latent action
$z_i\in\mathbb R^{d_z}$ for each transition $x_i\rightarrow x_{i+1}$.
We couple an IDM $q_\phi$, which infers latent actions from observed video, with an FDM $f_\theta$, which predicts their visual consequences:
\begin{equation}
    z_i
    \sim
    q_\phi
    \left(
        z_i
        \mid
        \widetilde{\mathbf x}^{\scriptscriptstyle\mathrm{IDM}}_{\le i+1},
        t_{\scriptscriptstyle\mathrm{IDM}}
    \right),
    \qquad
    \widehat{\mathbf v}
    =
    f_\theta
    \left(
        \widetilde{\mathbf x}^{\scriptscriptstyle\mathrm{FDM}},
        t_{\scriptscriptstyle\mathrm{FDM}},
        \mathbf c_0,
        \overline{\mathbf z}
    \right).
    \label{eq:two_dynamics}
\end{equation}
Here, $\mathbf c_0=\text{``a video''}$ is a fixed neutral text condition, and $\overline{\mathbf z} = [\,0,z_0,\ldots,z_{K-1}\,]$
aligns each $z_i$ with the prediction of $x_{i+1}$,
and $\widetilde{\mathbf x}^{\mathrm{IDM}}$ and
$\widetilde{\mathbf x}^{\mathrm{FDM}}$ denote the respective
inputs at independently sampled flow timesteps.
Both dynamics are initialized from the same pretrained video DiT~\cite{wan2025wan22, wan2025wan} and adapted independently with
LoRA~\cite{hu2022lora}. The new action-readout and action-conditioning modules
are trained from scratch.

\subsection{Video DiT as Inverse Dynamics}
\label{sec:video-dit-idm}

\input{fig/idm_attn}

The IDM repurposes the pretrained video DiT from an iterative generator into a
single-pass latent action encoder.

\noindent\textbf{Feed-forward repurposing.}
Unlike video generation, inverse dynamics operates on an observed video and requires no iterative denoising.
We therefore perform a single video DiT forward pass at an IDM flow timestep
$t_{\scriptscriptstyle\mathrm{IDM}}$~\citep{wang2026genception}.
The input is constructed as
$\widetilde{\mathbf x}^{\scriptscriptstyle\mathrm{IDM}} = (1-t_{\scriptscriptstyle\mathrm{IDM}})\mathbf x + t_{\scriptscriptstyle\mathrm{IDM}}\boldsymbol\epsilon.$
At $t_{\mathrm{IDM}}=0$, the model receives
clean video latents,
while increasing the timestep progressively removes visual information.
This provides a flexible family of noise-dependent representations from which to infer actions.

\noindent\textbf{Transition readout with action queries.}
To extract transition-level information, we introduce learnable action queries into the video DiT. A shared query embedding is replicated across transitions, with temporal
position encoding distinguishing each query $q_i$ associated with $x_i\rightarrow x_{i+1}$.
We employ an asymmetric attention mask
(Fig.~\ref{fig:idm_attention_mask}):
(i) video tokens attend causally across frames,
(ii) each action query $q_i$ reads only the two
endpoint states $(x_i,x_{i+1})$, and
(iii) action queries do not attend to one another,
nor do video tokens attend to the queries.
This makes each query a read-only transition register rather than an auxiliary communication channel.

\noindent\textbf{Variational action bottleneck.}
Let $h_i\in\mathbb R^D$ denote the final representation of action query $q_i$.
We use a linear Gaussian head
$g_\phi:\mathbb R^D\rightarrow\mathbb R^{2d_z}$, with $d_z\ll D$, to obtain a
compact latent action:
\begin{equation}
    [\,\mu_i,\log\sigma_i^2\,]
    =
    g_\phi\!\left(\operatorname{LN}(h_i)\right),
    \qquad
    q_\phi\!\left(
        z_i\mid
        \widetilde{\mathbf x}^{\mathrm{IDM}}_{\le i+1},
        t_{\mathrm{IDM}}
    \right)
    =
    \mathcal N\!\left(
        \mu_i,\operatorname{diag}(\sigma_i^2)
    \right).
    \label{eq:idm_readout}
\end{equation}
During training, we sample $z_i=\mu_i+\sigma_i\odot\xi_i$, where $\xi_i\sim\mathcal N(0,I)$ and use the posterior mean $\mu_i$ at inference.
The continuous latent space avoids quantization, allowing fine-grained variations in
visual dynamics to be represented~\citep{gaoadaworld, garrido2026learning, jiang2026olafworld}.

\input{fig/o2laf}

\subsection{Video DiT as Forward Dynamics}
\label{sec:video-dit-fdm}

The FDM predicts the visual consequences of inferred latent actions, providing the predictive supervision for the IDM.
To make this supervision informative, we adapt the pretrained
video DiT's temporal structure and conditioning to control information access, limiting visual shortcuts while preserving its generative capability.

\noindent\textbf{Initial-state and action conditioning.}
The FDM receives the clean initial state $x_0$ and the
temporally aligned latent-action sequence $\overline{\mathbf z}$.
The initial state provides visual context, including appearance and spatial structure, outside the latent-action bottleneck~\cite{ren2026videoworld2},
encouraging latent actions to capture transferable dynamics rather than scene-specific appearance.
Following standard DiT conditioning~\citep{peebles2023dit}, the aligned actions
are projected, combined with the corresponding flow-timestep embeddings, and injected into every Transformer block through AdaLN.

\noindent\textbf{Causal joint flow matching.}
We keep $x_0$ clean and add noise to all future states using a shared
timestep $t_{\scriptscriptstyle\mathrm{FDM}}$:
$
    \widetilde{\mathbf x}^{\scriptscriptstyle\mathrm{FDM}}
    =
    [\,
        x_0,
        x^{t_{\scriptscriptstyle\mathrm{FDM}}}_1,
        \ldots,
        x^{t_{\scriptscriptstyle\mathrm{FDM}}}_K
    \,],
    x^{t_{\scriptscriptstyle\mathrm{FDM}}}_i
    =
    (1-t_{\scriptscriptstyle\mathrm{FDM}})x_i
    +
    t_{\scriptscriptstyle\mathrm{FDM}}\epsilon_i.
    \label{eq:fdm_input}
$
The velocities of all future states are predicted jointly in a single forward pass at each flow timestep.
To prevent information leakage from later states, we replace the
pretrained video DiT's bidirectional temporal attention with
frame-causal attention, such that tokens at $x_i$ cannot attend
to later states $x_{>i}$.
When conditioned on clean preceding frames,
autoregressive prediction can exploit temporal redundancy
by simply copying the previous frame, bypassing the
latent-action bottleneck.
By removing this clean-history shortcut,
causal joint prediction encourages the FDM to leverage
latent actions, providing more effective supervision to the IDM.

\noindent\textbf{High-noise timestep sampling.}
Causal attention prevents future-state leakage, but each prediction still observes its own noisy target.
Common generative training recipes emphasize intermediate noise levels.
For example, SD3~\citep{esser2024scaling} uses logit-normal timestep sampling,
whose density vanishes at both endpoints.
While effective for visual generation, this middle-noise bias is undesirable for latent-action learning:
noisy future states may still retain sufficient information for the FDM to predict future states without fully utilizing latent actions.
We therefore emphasize high-noise
$t_{\scriptscriptstyle\mathrm{FDM}}$, reducing information available from the noisy targets
and encouraging the FDM to rely more strongly on latent actions.
The endpoint $t_{\scriptscriptstyle\mathrm{FDM}}=1$ makes this intuition
explicit.
For future state $x_i$, $v_i^\star=\epsilon_i-x_i$.
Defining $\hat x_i=\epsilon_i-\hat v_i$ gives
$
    \left\|
        \hat v_i-v_i^\star
    \right\|_2^2
    =
    \left\|
        \hat x_i-x_i
    \right\|_2^2.
    \label{eq:pure_noise_reconstruction}
$
Thus, at pure noise, flow matching is equivalent to reconstructing the future
latent conditioned on the clean initial state and latent actions.

\subsection{End-to-End Latent Action Learning}
\label{sec:end-to-end}

The IDM and FDM are connected through the sampled latent actions and optimized
end-to-end (Fig.~\ref{fig:pipeline}).

\noindent\textbf{Role-specific noise sampling.}
Noise plays different roles in the two dynamics.
For the IDM, it controls the visual information available for transition inference.
For the FDM, it controls how much target information remains accessible through the noisy future states.
We therefore sample their flow timesteps independently:
$
    t_{\scriptscriptstyle\mathrm{IDM}}
    \sim \pi_{\scriptscriptstyle\mathrm{IDM}},
    t_{\scriptscriptstyle\mathrm{FDM}}
    \sim \pi_{\scriptscriptstyle\mathrm{FDM}}.
    \label{eq:role_specific_timesteps}
$
The two distributions can thus be chosen for their respective roles rather
than inherited from the same generative training recipe.
In particular,
$\pi_{\scriptscriptstyle\mathrm{IDM}}=\delta_0$
corresponds to a clean-input IDM, while
$\pi_{\scriptscriptstyle\mathrm{FDM}}$
can emphasize high-noise states as described in
Sec.~\ref{sec:video-dit-fdm}.



\noindent\textbf{Learning objective.}
We optimize the flow-matching prediction loss~\cite{lipman2023flowmatching, esser2024scaling} together with a KL regularizer~\cite{kingma2014vae, higgins2017betavae}
on the latent-action posterior:
\begin{equation}
\begin{split}
    \mathcal L
    &=
    \mathbb E
    \!\left[
        \frac{1}{K}
        \sum_{i=1}^{K}
        w(t_{\scriptscriptstyle\mathrm{FDM}})
        \left\|
            \hat v_i-(\epsilon_i-x_i)
        \right\|_2^2
    \right] 
    +
    \frac{\beta}{K}
    \sum_{i=0}^{K-1}
    D_{\mathrm{KL}}\!\left[
        q_\phi(z_i\mid\cdot)
        \,\Vert\,
        \mathcal N(0,I)
    \right].
    \label{eq:joint_objective}
\end{split}
\end{equation}
Through reparameterization, gradients from the FDM propagate through the latent
actions into the IDM.
The prediction loss guides the IDM to encode transition
information essential for explaining future visual changes,
while the low-dimensional latent space and KL regularization
constrain the information carried by the action channel.

\noindent\textbf{Inference and action transfer.}
While the FDM shapes the latent-action space during training,
it serves as an action-conditioned video world model at inference.
Given an initial state and inferred latent actions, we initialize
future latents from Gaussian noise and integrate the learned flow
from $t=1$ to $t=0$ using the Euler solver~\citep{esser2024scaling}.
Actions inferred from a source video can thus condition generation from a different initial state, enabling zero-shot cross-context action transfer.

%% file: fig/motivation.tex
\begin{figure}[t]
  \centering
  \includegraphics[width=\linewidth]{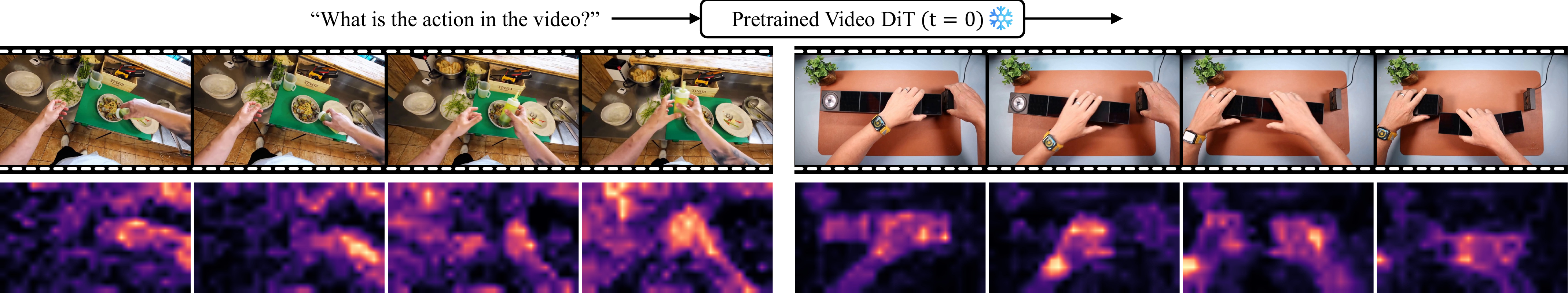}
  \vspace{-6mm}
  \caption{
    \textbf{Action-aware response in a pretrained video DiT.}
    Cross-attention follows the active hands and manipulated regions as the interaction changes.
    }
  \label{fig:motivation}
  \vspace{-3mm}
\end{figure}

%% file: fig/idm_attn.tex
\begin{wrapfigure}{r}{0.3\linewidth}
    \centering
    \vspace{-5mm}
    \includegraphics[width=\linewidth]{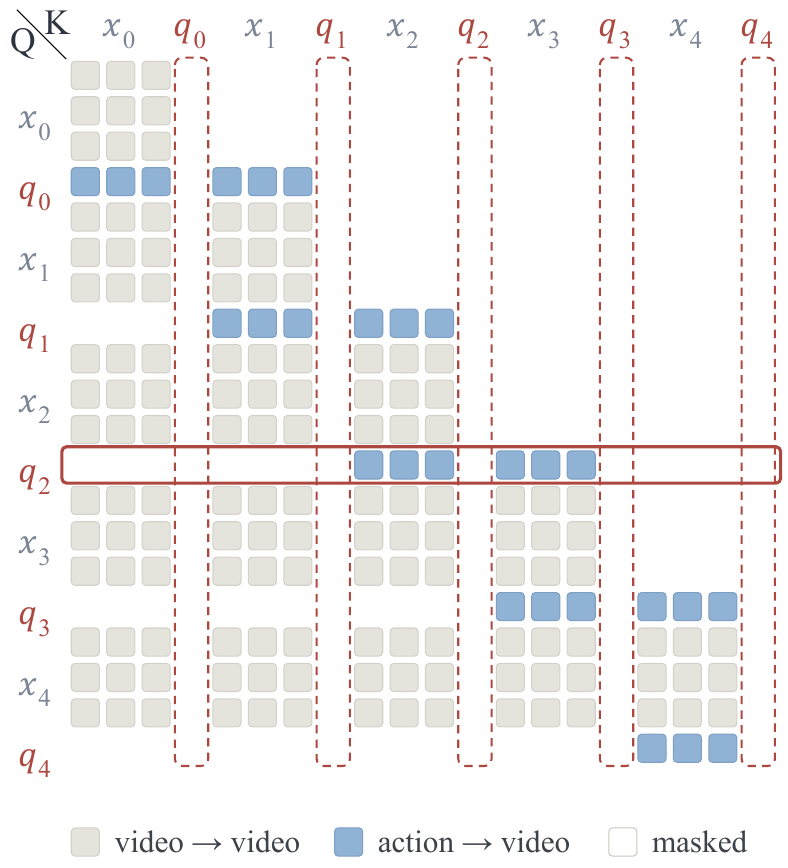}
    \captionsetup{
            font=footnotesize,
            labelfont=bf,
            skip=4pt
        }
    \caption{Action-query attention in IDM.}
    \label{fig:idm_attention_mask}
    \vspace{-5mm}
\end{wrapfigure}

%% file: fig/o2laf.tex
\begin{figure}[t]
  \centering
  \includegraphics[width=\linewidth]{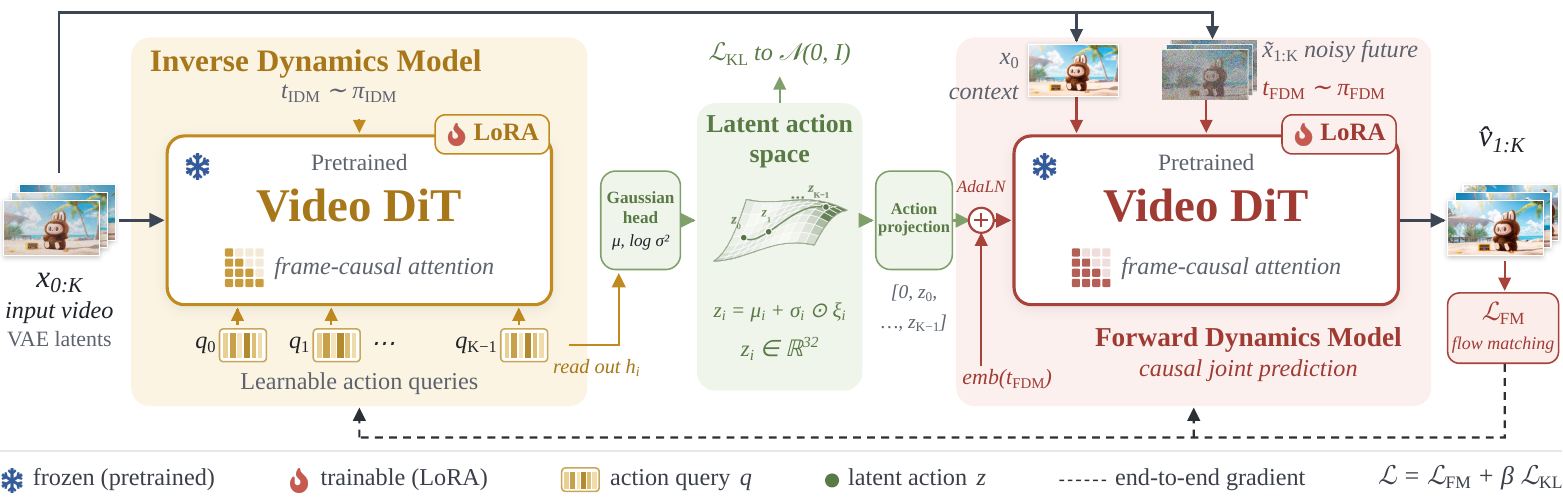}
  \caption{
\textbf{Overview of O$^{2}$LaF.}
Two independently adapted video DiTs share a pretrained prior:
the IDM infers latent actions, while the FDM predicts their
visual consequences.
Both employ frame-causal attention and role-specific
noise sampling, with end-to-end training through latent actions.
}
  \label{fig:pipeline}
  \vspace{-3mm}
\end{figure}

%% file: sec/4_exp.tex
\section{Experiments}
\label{sec:experiments}

We evaluate O$^{2}$LaF along two questions.
First, do the learned latent actions align with real controls across discrete
navigation and continuous manipulation
(Sec.~\ref{sec:exp:representation})?
Second, can the same actions serve as a unified control interface for video world model
in unseen visual contexts, without target-specific adaptation
(Sec.~\ref{sec:exp:transfer})?

\subsection{Experimental Setup}
\label{sec:exp:setup}

\noindent\textbf{Datasets.}
We evaluate O$^{2}$LaF across two domains with distinct visual dynamics.
For camera navigation, we train on MiraData~\cite{ju2024miradata} and evaluate on
MIND~\citep{ye2026mind}.
For robot manipulation, we train exclusively on egocentric human
demonstrations from EgoDex~\cite{hoque2026egodex} and directly evaluate on in-lab GR-1
robot demonstrations~\cite{gao2026dreamdojo}.
MiraData captures global scene and camera motion, whereas EgoDex
focuses on fine-grained hand--object interactions.
No action annotations are used to train the latent action models.

\noindent\textbf{Baselines.}
We compare with representative latent-action methods spanning the learning paradigms illustrated in Fig.~\ref{fig:paradigm}.
For navigation, we compare with AdaWorld~\cite{gaoadaworld} and Olaf-World~\cite{jiang2026olafworld}, all trained on the
same MiraData videos.
For manipulation, we compare with DreamDojo~\cite{gao2026dreamdojo}, VideoWorld2~\cite{ren2026videoworld2}, LAPA-DINOv3~\cite{ye2024lapa, nie2026lary}, and
VILLA-X~\cite{chen2026villax} using their released models.
Notably, O$^2$LaF uses only EgoDex subset of DreamDojo's large scale egocentric training data.
We evaluate representation quality using frozen encoders
under identical probing protocols, and  action-conditioned
generation through a controlled second-stage world-model comparison.

\noindent\textbf{Implementation details.}
Both the IDM and FDM of O$^{2}$LaF are initialized from
Wan2.2-TI2V-5B and adapted with rank-128 LoRA.
The IDM uses causal video-token attention, while the FDM uses frame-causal
attention with joint future prediction.
We use a 32-D Gaussian latent action with $\beta=10^{-4}$ and keep the video
VAE and text encoder frozen.
For all main experiments, we use a single domain-agnostic noise configuration:
$
    \pi_{\scriptscriptstyle\mathrm{IDM}}=\delta_0,
    \pi_{\scriptscriptstyle\mathrm{FDM}}
    =
    0.7\pi_{\mathrm{base}}
    +
    0.3\pi_{\mathrm{high}},
    \operatorname{supp}(\pi_{\mathrm{high}})
    \subseteq [0.9,1).
    \label{eq:canonical_training_noise}
$
Here, $\pi_{\mathrm{base}}$ is the standard shifted
flow-time distribution, with peak flow-matching weight
assigned to high-noise samples.
Sec.~\ref{sec:information_exposure} studies the broader
IDM--FDM noise design space across action domains.
For a fair downstream world-model evaluation, we train a separate bidirectional FDM for each
frozen latent-action encoder, using the same pretrained Wan2.2-TI2V-5B checkpoint and identical training and generation protocols, thereby isolating
the effect of latent-action representations.
Additional details are provided in Appendix~\ref{app:details}.

\input{fig/probing}


\subsection{Latent Action Quality}
\label{sec:exp:representation}

A useful latent action should make the underlying control easy to recover with
little labeled supervision.
We therefore freeze each latent-action encoder and fit lightweight linear
probes~\cite{zhang2022light} across increasing label budgets to evaluate
control decodability.

\noindent\textbf{Evaluation protocol.}
We evaluate discrete camera control and continuous robot manipulation.
On MIND, we predict four simultaneous ternary controls
(two translation and two rotation axes) using independent,
class-balanced logistic regression probes.
We report joint exact-match accuracy, requiring all four
predictions to be correct.
For manipulation, we use DreamDojo's In-lab Eval
benchmark~\cite{gao2026dreamdojo}, comprising GR-1
robot demonstrations.
We fit Ridge regression probes for wrist
translation and rotation on both arms.
We report EEF $R^2$, averaged over translation,
rotation, and both arms.
Label budgets range from 1 to 50 videos for navigation
and 1 to 32 episodes per task for manipulation.

\noindent\textbf{Results.}
Fig.~\ref{fig:latent_action_fewshot} shows that
O$^{2}$LaF consistently improves linear decodability
across domains and label budgets.
Performance rises sharply with the first few
labeled examples, indicating that control-relevant information is readily accessible with limited supervision.
Notably, O$^{2}$LaF achieves an EEF $R^2$ of $0.277$
on GR-1, compared with $0.149$ for DreamDojo,
despite training its encoder exclusively on human
demonstrations without robot-domain adaptation, suggesting that the learned latent actions capture motion structure transferable across human and robot embodiments.
Together, these results demonstrate that O$^{2}$LaF
learns latent actions that admit label-efficient mappings
to both discrete navigation controls and continuous robot motions.


\subsection{Zero-Shot Cross-Context Action Transfer}
\label{sec:exp:transfer}

Beyond linear decodability, we investigate whether
latent actions provide a transferable control interface
across visual contexts and embodiments.
Given a reference video, we extract its latent-action sequence
and condition the FDM on a different target first frame,
without any adaptation.

\subsubsection{Camera navigation}

\noindent\textbf{Evaluation protocol.} 
We extract actions from 24 reference trajectories
(12 first-person and 12 third-person) and transfer them
to 50 and 20 novel visual contexts, respectively,
yielding 840 cross-context generations.
All methods share identical references, target contexts,
and generation settings.
We assess camera-motion fidelity by estimating trajectories
from reference and generated videos using
ViPE~\cite{huang2025vipe}, reporting Sim(3)-aligned
translation NRMSE and rotation RPE.
We separately evaluate visual quality using
VBench~\cite{huang2024vbench}.

\noindent\textbf{Results.}
Tab.~\ref{tab:world_model_comparison} shows that
O$^{2}$LaF achieves the lowest camera-motion errors
and highest imaging quality in both perspectives.
Interestingly, latent-action CFG consistently improves
camera-motion fidelity and imaging quality, resembling
the effectiveness of text-based CFG~\cite{ho2022cfg}.
This contrasts with prior findings that CFG is less
effective for predefined action controls~\cite{rigter2024avid}.
We hypothesize that the learned latent actions form a
meaningful continuous motion manifold, enabling CFG to
amplify motion-relevant conditioning and improve sample
fidelity.
Qualitatively, Fig.~\ref{fig:actiontransfermind} illustrates the transfer of reference motion across distinct visual contexts.
In first-person navigation, O$^{2}$LaF reproduces
the reference camera's progressive upward reorientation,
while baselines exhibit limited motion or inaccurate viewpoint changes.
In third-person navigation, O$^{2}$LaF captures
the agent's rightward turn and leaning posture,
together with corresponding background motion,
whereas baselines largely fail to reproduce the turn.

\input{table/fdm_mira}
\input{fig/fdm_transfer_mind}

\input{table/fdm_egodex}
\input{fig/fdm_transfer_robot}

\subsubsection{Robot manipulation}

\noindent\textbf{Evaluation protocol.}
We evaluate human-to-robot and robot-to-human action transfer
using 85 video pairs spanning 20 manipulation tasks.
Human videos are drawn from EgoDex, while robot videos come from
DreamDojo's EgoDex Eval benchmark~\cite{gao2026dreamdojo},
which reproduces EgoDex interactions using the GR-1 robot.
Since the task-paired human and robot demonstrations
are neither temporally nor spatially aligned,
we evaluate direct action transfer qualitatively.
For quantitative evaluation, following~\citet{garrido2026learning},
we measure self-reconstruction and cross-embodiment cycle consistency
($H\!\rightarrow\!R\!\rightarrow\!H$ and
$R\!\rightarrow\!H\!\rightarrow\!R$)
using PSNR, SSIM, and LPIPS.

\noindent\textbf{Results.}
Tab.~\ref{tab:robot_crsos_transfer} shows that O$^{2}$LaF
outperforms DreamDojo across all self-reconstruction
and cross-cycle metrics, suggesting that the learned actions
retain motion information across embodiment.
Qualitatively, Fig.~\ref{fig:actiontransferrobot}
illustrates that human reaching motions induce corresponding
robot arm movements, while robot actions elicit human
reaching and object-directed interactions.
Notably, O$^{2}$LaF preserves arm-specific motion,
including the active arm and reaching direction,
despite substantial embodiment differences.
Compared with DreamDojo, it exhibits more faithful
motion correspondence while preserving the target
visual context.

%% file: fig/probing.tex
\begin{figure*}[t]
    \centering
    \begin{minipage}[t]{0.495\textwidth}
        \centering
        \includegraphics[
            width=\linewidth
        ]{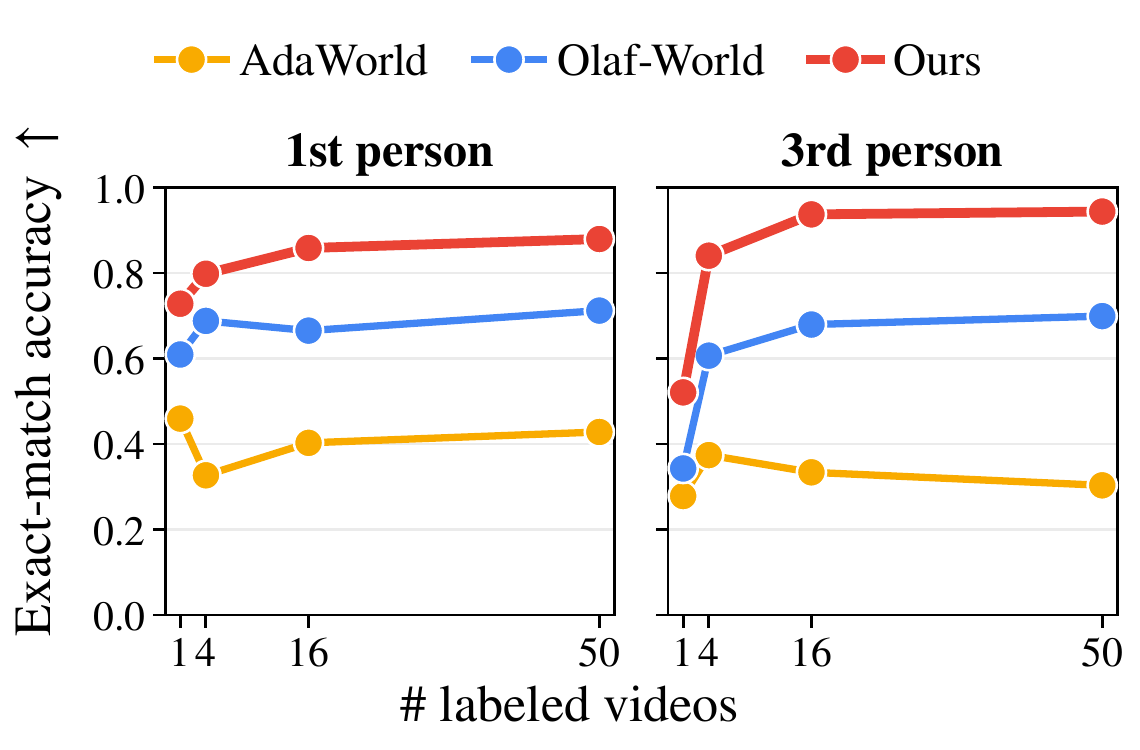}
        \vspace{-3mm}

        {\small\textbf{(a) Camera navigation on MIND.}}
    \end{minipage}
    \hfill
    \begin{minipage}[t]{0.495\textwidth}
        \centering
        \includegraphics[
            width=\linewidth
        ]{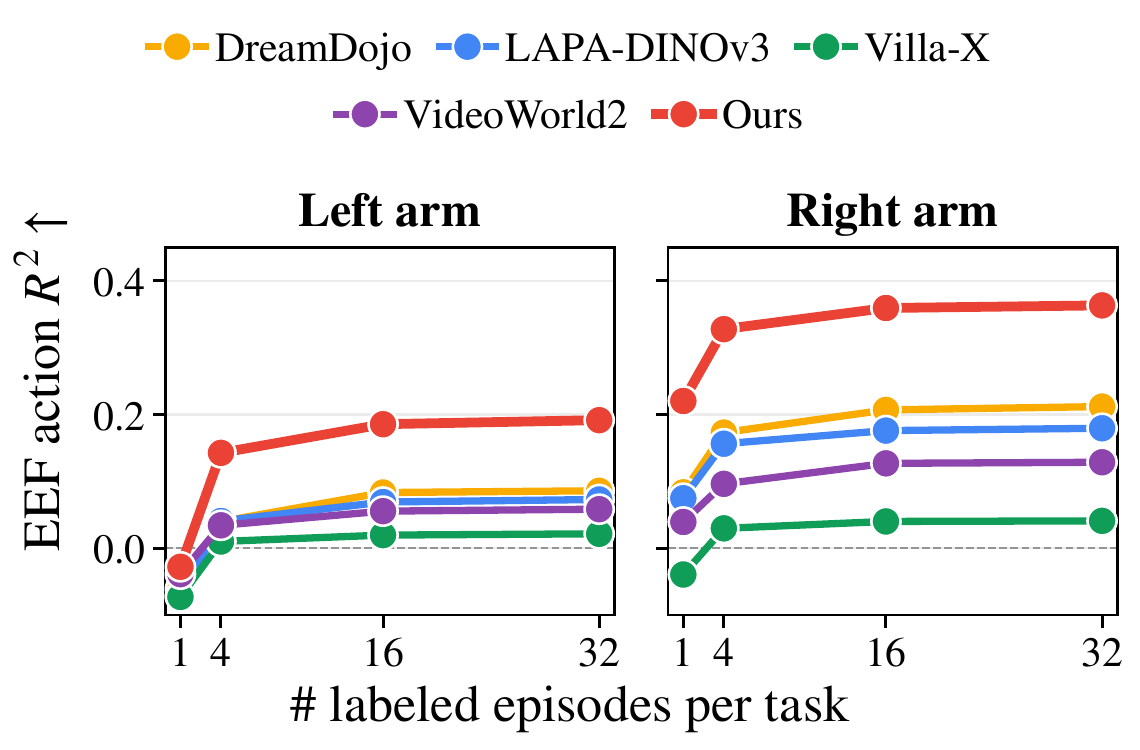}
        \vspace{-3mm}

        {\small\textbf{(b) Robot manipulation on In-lab EEF.}}
    \end{minipage}
    \vspace{-0.4em}

    \caption{\textbf{Few-shot linear probing of frozen latent actions.}
    O$^2$LaF consistently outperforms prior latent-action representations on
    both settings across labeled-data budgets.
    }
    \label{fig:latent_action_fewshot}
    \vspace{-3mm}
\end{figure*}

%% file: table/fdm_mira.tex
\begin{table*}[t]
\centering
\caption{
\textbf{Zero-shot cross-context camera navigation.}
We report VBench imaging quality (Img. Q.) and temporal
consistency (Temp. C.), and ViPE-based
translation NRMSE and rotation RPE.
Methods above the separator use action CFG$=3$.
Best and second-best results are \textbf{bold} and
\underline{underlined}, respectively.
}
\vspace{-2mm}
\label{tab:world_model_comparison}

\small
\setlength{\tabcolsep}{4.5pt}
\setlength{\aboverulesep}{0pt}
\setlength{\belowrulesep}{0pt}
\renewcommand{\arraystretch}{1.05}
\setlength{\extrarowheight}{0.5pt}

\begin{tabular}{l|cccc|cccc}
\toprule
\multirow{3}{*}{\textbf{Method}}
& \multicolumn{4}{c|}{\textbf{\textsc{1st-P}}}
& \multicolumn{4}{c}{\textbf{\textsc{3rd-P}}} \\
\cmidrule(lr){2-5} \cmidrule(lr){6-9}

& \multicolumn{2}{c}{\textbf{Visual Quality}}
& \multicolumn{2}{c|}{\textbf{Camera Motion}}
& \multicolumn{2}{c}{\textbf{Visual Quality}}
& \multicolumn{2}{c}{\textbf{Camera Motion}} \\
\cmidrule(lr){2-3} \cmidrule(lr){4-5}
\cmidrule(lr){6-7} \cmidrule(lr){8-9}

& Img.~Q. $\uparrow$
& Temp.~C. $\uparrow$
& Trans. $\downarrow$
& Rot. $\downarrow$
& Img.~Q. $\uparrow$
& Temp.~C. $\uparrow$
& Trans. $\downarrow$
& Rot. $\downarrow$ \\
\midrule

AdaWorld~\cite{gaoadaworld}
& 0.639 & 0.899 & 2.539 & 2.055
& 0.616 & \underline{0.927} & 1.342 & 1.217 \\

Olaf-World~\cite{jiang2026olafworld}
& \underline{0.690} & \textbf{0.910} & 1.627 & 1.300
& 0.648 & \textbf{0.930} & 0.985 & 0.701 \\

Ours
& \textbf{0.698} & \underline{0.900} & \textbf{1.289} & \textbf{0.786}
& \textbf{0.682} & 0.919 & \textbf{0.511} & \textbf{0.330} \\

\midrule

Ours (w/o CFG)
& 0.672 & \textbf{0.910} & \underline{1.364} & \underline{0.830}
& \underline{0.662} & \underline{0.927} & \underline{0.513} & \underline{0.364} \\

\bottomrule
\end{tabular}

\vspace{-2mm}
\end{table*}

%% file: fig/fdm_transfer_mind.tex
\begin{figure*}[t]
\includegraphics[width=\linewidth]{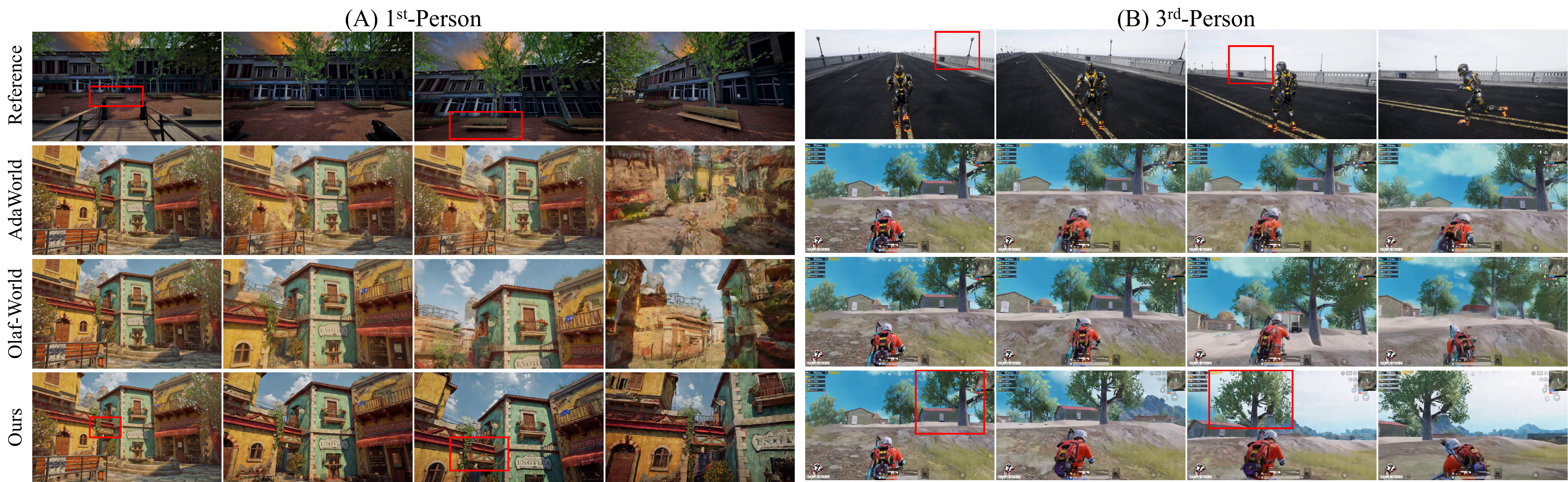}
\vspace{-6mm}
\caption{
\textbf{Zero-shot cross-context camera navigation.}
O$^{2}$LaF better reproduces reference camera and agent
action while preserving the target scene.
}
\vspace{-4mm}
\label{fig:actiontransfermind}
\end{figure*}

%% file: table/fdm_egodex.tex
\begin{table*}[t]
  \centering
  \caption{
  \textbf{Cross-embodiment action transfer via cycle reconstruction.}
  \textit{Self} reconstructs the source video directly, while
  \textit{Cross-cycle} reconstructs it after transferring latent
  actions to the opposite embodiment and back ($H\!\rightarrow\!R\!\rightarrow\!H$ and $R\!\rightarrow\!H\!\rightarrow\!R$).
  Best results are in \textbf{bold}.
  }
  \label{tab:robot_crsos_transfer}
  \vspace{-2mm}

  \footnotesize
  \setlength{\tabcolsep}{3.2pt}
  \setlength{\aboverulesep}{0pt}
  \setlength{\belowrulesep}{0pt}
  \renewcommand{\arraystretch}{1.05}
  \setlength{\extrarowheight}{0.5pt}

  \begin{tabular}{l|cccccc|cccccc}
  \toprule
  \multirow{3}{*}{\textbf{Method}}
  & \multicolumn{6}{c|}{\textbf{\textsc{Human}}}
  & \multicolumn{6}{c}{\textbf{\textsc{Robot}}} \\
  \cmidrule(lr){2-7} \cmidrule(lr){8-13}

  & \multicolumn{3}{c}{\textbf{Self}}
  & \multicolumn{3}{c|}{\textbf{Cross-cycle}}
  & \multicolumn{3}{c}{\textbf{Self}}
  & \multicolumn{3}{c}{\textbf{Cross-cycle}} \\
  \cmidrule(lr){2-4} \cmidrule(lr){5-7}
  \cmidrule(lr){8-10} \cmidrule(lr){11-13}

  & PSNR$\uparrow$ & SSIM$\uparrow$ & LPIPS$\downarrow$
  & PSNR$\uparrow$ & SSIM$\uparrow$ & LPIPS$\downarrow$
  & PSNR$\uparrow$ & SSIM$\uparrow$ & LPIPS$\downarrow$
  & PSNR$\uparrow$ & SSIM$\uparrow$ & LPIPS$\downarrow$ \\
  \midrule

  DreamDojo~\cite{gao2026dreamdojo}
  & 23.38 & 0.871 & 0.223
  & 22.61 & 0.859 & 0.244
  & 18.73 & 0.769 & 0.159
  & 18.35 & 0.754 & 0.165 \\

  Ours
  & \textbf{26.72} & \textbf{0.903} & \textbf{0.148}
  & \textbf{24.30} & \textbf{0.878} & \textbf{0.195}
  & \textbf{19.92} & \textbf{0.787} & \textbf{0.140}
  & \textbf{18.85} & \textbf{0.770} & \textbf{0.163} \\

  \bottomrule
  \end{tabular}

  \vspace{-2mm}
\end{table*}

%% file: fig/fdm_transfer_robot.tex
\begin{figure*}[t]
\includegraphics[width=\linewidth]{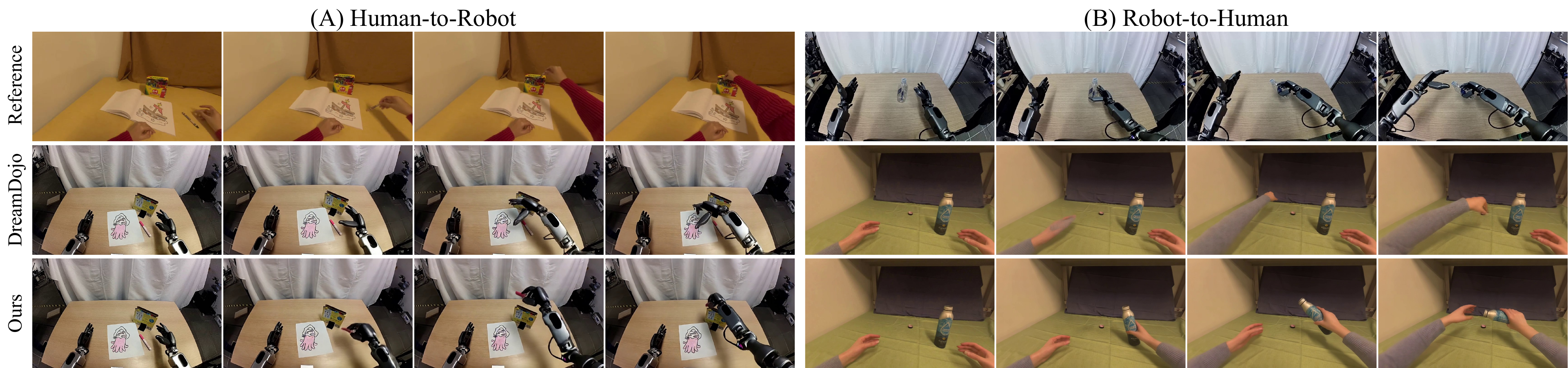}
\caption{
\textbf{Zero-shot cross-embodiment manipulation}
O$^{2}$LaF better preserves object-directed motion
while adapting to the target embodiment and visual context.
}
\vspace{-3mm}
\label{fig:actiontransferrobot}
\end{figure*}

%% file: sec/5_analysis.tex
\input{table/elucidating_1}
\input{fig/noise_sweep}

\section{Understanding the Two Dynamics}
\label{sec:understanding}

Although sharing the same pretrained video prior, the IDM and FDM serve distinct roles in latent action learning.
We systematically investigate how their design choices shape the learned action representations and their utility for controllable video generation.

\subsection{Adapting Pretrained Dynamics}
\label{sec:capacity}

Pretrained video DiTs encode rich spatiotemporal knowledge,
but their representations are not explicitly optimized for
action inference.
We investigate whether this knowledge can be extracted
through a lightweight action head or requires adapting
the pretrained backbone.
Since the IDM learns through predictive supervision from
the FDM, we further study whether a stronger FDM
improves the learned action representations.

Tab.~\ref{tab:idm_adaptation} shows that a frozen video DiT
with a trainable action head performs substantially worse
than LoRA adaptation, while rank-16 LoRA approaches rank-128
performance.
Meanwhile, scaling the FDM from 1.3B to 14B consistently improves action
decodability with a fixed-size IDM (Tab.~\ref{tab:fdm_scaling}).
These results suggest that effective latent-action learning
benefits from both adapting pretrained representations and
strengthening predictive supervision.


\subsection{Controlling Information Exposure}
\label{sec:information_exposure}

We revisit how the information-access designs of our two
dynamics shape latent-action learning.

\noindent\textbf{Temporal factorization.}
We investigate how the FDM's temporal prediction structure shapes the learned actions.
As shown in Tab.~\ref{tab:temporal_factorization},
causal joint prediction substantially outperforms
autoregression across both navigation perspectives
and achieves the strongest overall probing performance.

\noindent\textbf{Role-specific noise levels.}
Motivated by the view of diffusion noise as partial
masking~\citep{chen2024diffusion}, we investigate how
noise exposure in the two dynamics shape latent-action learning.
We conduct a $5\times5$ sweep over fixed
$(t_{\mathrm{IDM}},t_{\mathrm{FDM}})$ pairs
across navigation and manipulation.
Fig.~\ref{fig:noise_sweep} reveals three findings.
First, moderate IDM noise often improves action decodability
over clean inputs, potentially favoring higher-level,
action-relevant features over appearance-specific details~\citep{tang2023emergent}.
Second, low FDM noise yields poor representations,
consistent with shortcuts through partially observed targets.
However, maximal noise is not always optimal, potentially due to high-variance velocity targets near the noise prior~\citep{yang2026stablevelocity},
which weakens predictive supervision for the IDM.
Finally, preferred noise configurations vary across domains:
both navigation perspectives favor
$(t_{\mathrm{IDM}},t_{\mathrm{FDM}})=(0.3,0.98)$,
whereas manipulation exhibits different optima across arms.
%


\input{table/elucidating_idm_priors}

\input{fig/prior_fdm_transfer}

\subsection{Extending to Other Video Priors}
\label{sec:idm_prior}

Our inverse-forward dynamics framework extends beyond generative video DiTs to other video representations. 
We explore four IDM backbones spanning distinct learning
paradigms:
(i) ST-Transformer~\cite{bruce2024genie}, a spatiotemporal Transformer trained from scratch,
(ii) VideoChat3~\cite{li2026videochat3}, a language-aligned video MLLM,
(iii) V-JEPA2~\cite{assran2025vjepa2}, a self-supervised
predictive video encoder, and (iv) Wan2.2~\cite{wan2025wan22}, a generative video DiT.
We similarly adapt each backbone as an IDM by introducing causal temporal attention and learnable action queries, while retaining the same FDM design.

Tab.~\ref{tab:idm_prior} reveals distinct behaviors.
Notably, V-JEPA2 achieves the strongest action probing performance, 
yet its latent actions exhibit noticeable source-appearance
leakage during cross-context generation
(Fig.~\ref{fig:actiontransfer_idmprior}).
Color probing further confirms substantially higher
appearance decodability in V-JEPA2 latent actions,
whereas Wan2.2 better preserves the target scene.
We hypothesize that sharing a generative video prior
between the IDM and FDM improves their representation
alignment, facilitating more transferable generative control.
It highlights that action probing alone
does not reveal whether latent actions also encode
task-irrelevant information that may hinder transfer.
Downstream generative evaluation is therefore essential
for assessing the transferability of learned action representations.

\input{table/elucidating_fdm_video}

\subsection{Balancing Action Learning and Generation}
\label{sec:deployment}

Jointly learning latent actions and a generative
video world model is a compelling objective~\citep{wang2026coevolving}.
Yet the FDM faces a practical trade-off: learning informative
actions requires limiting visual shortcuts, whereas
high-fidelity generation benefits from broader noise coverage.
In O$^{2}$LaF, causal prediction and high-noise supervision
serve the former objective, but compromise the latter.
We investigate whether both objectives can be reconciled
within end-to-end training.

We first study FDM noise sampling and IDM gradient routing.
Tab.~\ref{tab:learning_deployment} shows that mixed noise
sampling substantially improves generation over
high-noise-only training while maintaining comparable
action probing performance.
However, restricting IDM gradients to high-noise timesteps
provides no consistent probing gains.
Given the inherent tension, we adopt a two-stage approach~\cite{jiang2026olafworld}, freezing the learned
IDM and training a bidirectional FDM.
This substantially improves generation quality.
Moreover, initializing from the jointly trained FDM,
rather than the original pretrained video DiT,
further improves both visual quality and action-following
fidelity.

%% file: table/elucidating_1.tex
\begin{table*}[b]
\centering
\vspace{-3mm}
\caption{
\textbf{Design choices for the two dynamics.}
MIND joint exact-match accuracy ($\%$, $\uparrow$)
at $k\in\{1,50\}$.
Panels (a,c) use Wan2.2-5B, while (b) studies
Wan2.1 FDM scaling with a fixed 1.3B IDM.
}
\vspace{-2mm}
\label{tab:design_space_summary}

\scriptsize
\renewcommand{\arraystretch}{1.12}
\setlength{\tabcolsep}{3.2pt}
\captionsetup[subtable]{
    font=footnotesize,
    labelfont=normalfont,
    textfont=bf,
    skip=3pt
}

\sbox{\panelbox}{%
\begin{tabular}{@{}l|cc|cc@{}}
\toprule
\multirow{2}{*}{\textbf{Variant}}
& \multicolumn{2}{c|}{\textbf{\textsc{1st-P}}}
& \multicolumn{2}{c}{\textbf{\textsc{3rd-P}}} \\
\cmidrule(lr){2-3}\cmidrule(lr){4-5}
& $k{=}1$ & $k{=}50$ & $k{=}1$ & $k{=}50$ \\
\midrule
Head-only
& 43.88 & 39.33
& 22.15 & 44.27 \\
LoRA-16
& \underline{70.27} & \underline{84.73}
& \textbf{55.12} & \underline{93.92} \\
LoRA-128
& \textbf{72.81} & \textbf{87.95}
& \underline{52.08} & \textbf{94.37} \\
\bottomrule
\end{tabular}}
\begin{subtable}[t]{\wd\panelbox}
\centering
\caption{IDM Adaptation}
\label{tab:idm_adaptation}
\usebox{\panelbox}
\end{subtable}
\hfill
\sbox{\panelbox}{%
\begin{tabular}{@{}l|cc|cc@{}}
\toprule
\multirow{2}{*}{\textbf{IDM/FDM}}
& \multicolumn{2}{c|}{\textbf{\textsc{1st-P}}}
& \multicolumn{2}{c}{\textbf{\textsc{3rd-P}}} \\
\cmidrule(lr){2-3}\cmidrule(lr){4-5}
& $k{=}1$ & $k{=}50$ & $k{=}1$ & $k{=}50$ \\
\midrule
1.3B/1.3B
& 67.14 & 80.76
& \underline{59.68} & 89.52 \\
1.3B/14B
& \underline{72.72} & \underline{85.98}
& \textbf{61.73} & \underline{92.19} \\
5B/5B (ref)
& \textbf{72.81} & \textbf{87.95}
& 52.08 & \textbf{94.37} \\
\bottomrule
\end{tabular}}
\begin{subtable}[t]{\wd\panelbox}
\centering
\caption{FDM Scale}
\label{tab:fdm_scaling}
\usebox{\panelbox}
\end{subtable}
\hfill
\sbox{\panelbox}{%
\begin{tabular}{@{}l|cc|cc@{}}
\toprule
\multirow{2}{*}{\textbf{Variant}}
& \multicolumn{2}{c|}{\textbf{\textsc{1st-P}}}
& \multicolumn{2}{c}{\textbf{\textsc{3rd-P}}} \\
\cmidrule(lr){2-3}\cmidrule(lr){4-5}
& $k{=}1$ & $k{=}50$ & $k{=}1$ & $k{=}50$ \\
\midrule
Bidir. joint
& \textbf{72.90} & \underline{83.71}
& 45.54 & \underline{92.36} \\
Causal joint
& \underline{72.81} & \textbf{87.95}
& \textbf{52.08} & \textbf{94.37} \\
AR ($h{=}1$)
& 55.45 & 70.13
& \underline{45.79} & 79.49 \\
\bottomrule
\end{tabular}}
\begin{subtable}[t]{\wd\panelbox}
\centering
\caption{FDM Factorization}
\label{tab:temporal_factorization}
\usebox{\panelbox}
\end{subtable}

\vspace{-2mm}
\end{table*}

%% file: fig/noise_sweep.tex
\begin{figure*}[h]
    \centering
    \begin{minipage}[t]{0.495\textwidth}
        \centering
        \includegraphics[
            width=\linewidth
        ]{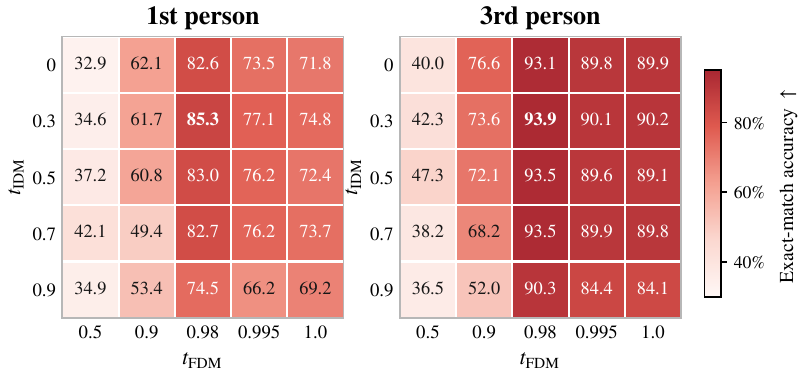}
        \vspace{-1.2em}

        {\small\textbf{(a) Navigation (MIND accuracy $\uparrow$, $k=50$).}}
    \end{minipage}
    \hfill
    \begin{minipage}[t]{0.495\textwidth}
        \centering
        \includegraphics[
            width=\linewidth
        ]{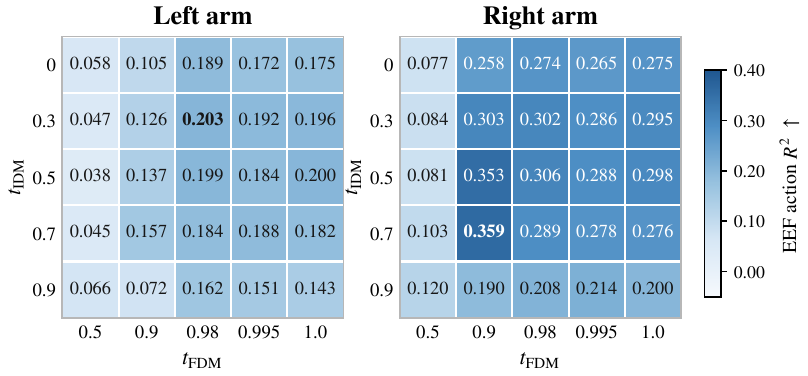}
        \vspace{-1.2em}

        {\small\textbf{(b) Manipulation (EEF $R^2 \uparrow$, $k=32$).}}
    \end{minipage}
    \vspace{-0.4em}

    \caption{
\textbf{Role-specific noise levels.}
IDM linear probing across
a $5\times5$ grid of IDM and FDM flow times.
The non-monotonic trends reveal distinct preferred
noise configurations across motion domains.
}
    \label{fig:noise_sweep}
    \vspace{-5mm}
\end{figure*}

%% file: table/elucidating_idm_priors.tex
\begin{table*}[!b]
\centering
\caption{
\textbf{Comparison of O$^2$LaF IDM backbones.}
Color $R^2$ measures the linear decodability of mean
reference-frame CIELAB ($L/a/b$) from latent actions.
Positive values indicate appearance leakage, while
non-positive values indicate no improvement over the mean predictor.
}
\vspace{-3mm}
\label{tab:idm_prior}

\scriptsize
\setlength{\tabcolsep}{2.5pt}
\setlength{\aboverulesep}{0pt}
\setlength{\belowrulesep}{0pt}
\renewcommand{\arraystretch}{1.08}
\newcolumntype{Y}{>{\centering\arraybackslash}X}

\begin{tabularx}{\textwidth}{@{}l|YY|c|YYYY|YYYY@{}}
\toprule
\multirow{3}{*}{\textbf{IDM Backbone}}
& \multicolumn{3}{c|}{\textbf{Probing}}
& \multicolumn{4}{c|}{\textbf{\textsc{1st-P} World Model}}
& \multicolumn{4}{c}{\textbf{\textsc{3rd-P} World Model}} \\
\cmidrule(lr){2-4}
\cmidrule(lr){5-8}
\cmidrule(lr){9-12}

& \multicolumn{2}{c|}{\textbf{MIND} $\uparrow$}
& \multirow{2}{*}{
    \shortstack{\textbf{Color $R^2$}\\
    \textbf{$L$ / $a$ / $b$}}
}
& \multicolumn{2}{c}{\textbf{VBench} $\uparrow$}
& \multicolumn{2}{c|}{\textbf{ViPE} $\downarrow$}
& \multicolumn{2}{c}{\textbf{VBench} $\uparrow$}
& \multicolumn{2}{c}{\textbf{ViPE} $\downarrow$} \\
\cmidrule(lr){2-3}
\cmidrule(lr){5-6}
\cmidrule(lr){7-8}
\cmidrule(lr){9-10}
\cmidrule(lr){11-12}

& 1st-P & 3rd-P
&
& Img. & Temp. & Trans. & Rot.
& Img. & Temp. & Trans. & Rot. \\
\midrule

ST-Transformer
& 58.79 & 79.70
& $-0.419\,/\,{-}0.190\,/\,{-}0.530$
& \underline{0.666} & \textbf{0.901} & 1.449 & 1.022
& 0.636 & \textbf{0.924} & 0.620 & 0.478 \\

VideoChat3
& 36.96 & 43.86
& $0.411\,/\,{-}0.147\,/\,0.364$
& -- & -- & -- & --
& -- & -- & -- & -- \\

V-JEPA2
& \textbf{88.62} & \textbf{96.14}
& $0.514\,/\,0.816\,/\,0.742$
& 0.657 & \textbf{0.901} & \underline{1.349} & \underline{0.847}
& \underline{0.664} & 0.916 & \textbf{0.439} & \underline{0.337} \\

\midrule

\rowcolor{black!5}
Wan2.2
& \underline{87.95} & \underline{94.37}
& $-0.297\,/\,0.081\,/\,0.223$
& \textbf{0.698} & \underline{0.900} & \textbf{1.289} & \textbf{0.786}
& \textbf{0.682} & \underline{0.919} & \underline{0.511} & \textbf{0.330} \\

\bottomrule
\end{tabularx}

\vspace{-2mm}
\end{table*}

%% file: fig/prior_fdm_transfer.tex
\begin{figure*}[!b]
\includegraphics[width=\linewidth]{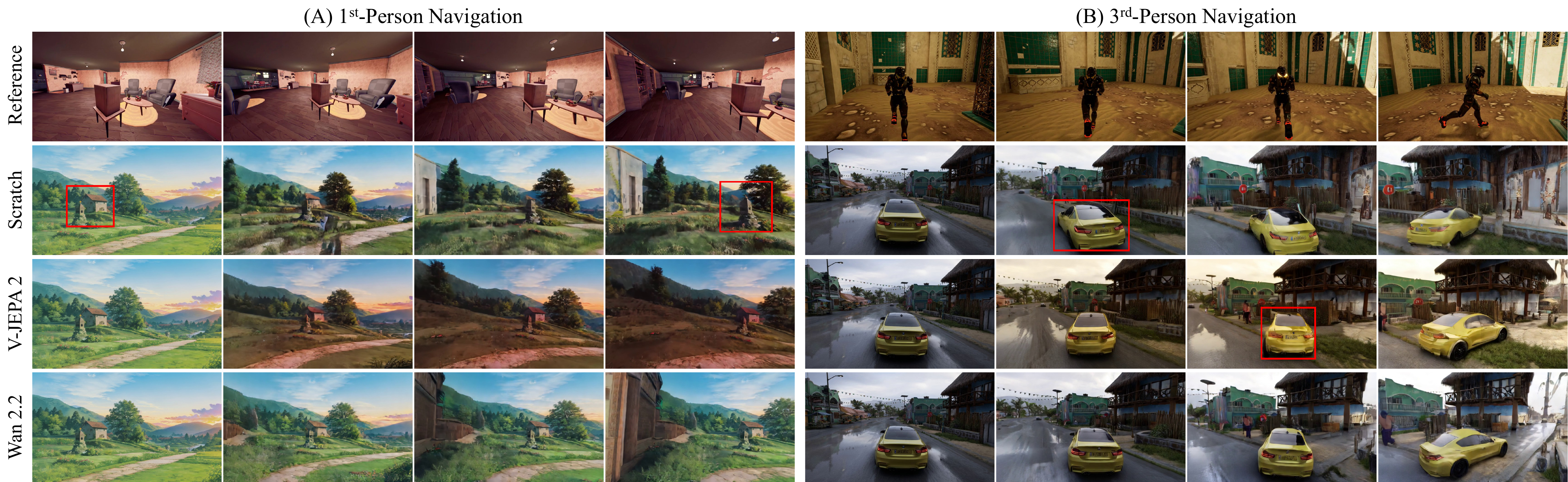}
\vspace{-5mm}
\caption{
\textbf{Cross-context action transfer with IDM priors.}
V-JEPA2 exhibits source-appearance leakage,
whereas video DiT better preserves target appearance.
}
\vspace{-3mm}
\label{fig:actiontransfer_idmprior}
\end{figure*}

%% file: table/elucidating_fdm_video.tex
\begin{table*}[t]
\centering
\caption{
\textbf{Trade-offs between action learning and generation.}
End-to-end models vary FDM noise sampling
(\textit{Mixed}: standard and high; \textit{High}: high only)
and IDM gradient routing
(\textit{All}: all timesteps; \textit{High}: high-noise only).
Two-stage models freeze the \textit{Mixed/All} IDM
and compare pretrained versus E2E FDM initialization.
Gray probing scores are shared and excluded from ranking.
}
\vspace{-3mm}
\label{tab:learning_deployment}

\footnotesize
\setlength{\tabcolsep}{4.25pt}
\setlength{\aboverulesep}{0pt}
\setlength{\belowrulesep}{0pt}
\renewcommand{\arraystretch}{1.05}
\setlength{\extrarowheight}{0.5pt}

\begin{tabular}{ll!{\vrule}cccccc!{\vrule}cccccc}
\toprule
\multicolumn{2}{c!{\vrule}}{\textbf{Configuration}}
& \multicolumn{6}{c|}{\textbf{\textsc{1st-P}}}
& \multicolumn{6}{c}{\textbf{\textsc{3rd-P}}} \\
\cmidrule(lr){3-8} \cmidrule(lr){9-14}

& & \multicolumn{2}{c}{\textbf{Probing}$\uparrow$}
& \multicolumn{2}{c}{\textbf{VBench} $\uparrow$}
& \multicolumn{2}{c|}{\textbf{ViPE} $\downarrow$}
& \multicolumn{2}{c}{\textbf{Probing}$\uparrow$}
& \multicolumn{2}{c}{\textbf{VBench} $\uparrow$}
& \multicolumn{2}{c}{\textbf{ViPE} $\downarrow$} \\
\cmidrule(lr){3-4} \cmidrule(lr){5-6} \cmidrule(lr){7-8}
\cmidrule(lr){9-10} \cmidrule(lr){11-12} \cmidrule(lr){13-14}

FDM noise & IDM grad. & $k{=}1$ & $k{=}50$
& Img. & Temp.
& Trans. & Rot.
& $k{=}1$ & $k{=}50$
& Img. & Temp.
& Trans. & Rot.\\
\midrule

\multicolumn{14}{l}{\textit{End-to-end learning (causal FDM)}} \\

\rowcolor{themecolor!8}
Mixed & All
& \textbf{72.81} & \textbf{87.95}
& 0.649 & 0.895 & 1.937 & 0.993
& 52.08 & 94.37
& 0.654 & \underline{0.913} & 0.845 & 0.481 \\

Mixed & High
& 72.23 & \underline{86.79}
& 0.654 & 0.885 & 1.959 & 0.943
& \underline{60.38} & \textbf{95.11}
& 0.660 & 0.902 & 0.918 & 0.484 \\

High & High
& \underline{72.28} & 85.76
& 0.647 & 0.836 & 2.419 & 1.168
& \textbf{60.75} & \underline{94.86}
& 0.646 & 0.853 & 1.618 & 0.765 \\

\midrule

\multicolumn{14}{l}{\textit{Two-stage deployment (bidirectional FDM)}} \\

\rowcolor{themecolor!8}
\multicolumn{2}{l!{\vrule}}{Pretrained init.}
& \textcolor{black!45}{72.81} & \textcolor{black!45}{87.95}
& \underline{0.698} & \underline{0.900} & \underline{1.289} & \underline{0.786}
& \textcolor{black!45}{52.08} & \textcolor{black!45}{94.37}
& \underline{0.682} & \textbf{0.919} & \underline{0.511} & \underline{0.330} \\

\rowcolor{themecolor!8}
\multicolumn{2}{l!{\vrule}}{E2E init.}
& \textcolor{black!45}{72.81} & \textcolor{black!45}{87.95}
& \textbf{0.711} & \textbf{0.901} & \textbf{1.260} & \textbf{0.725}
& \textcolor{black!45}{52.08} & \textcolor{black!45}{94.37}
& \textbf{0.686} & \textbf{0.919} & \textbf{0.459} & \textbf{0.322} \\

\bottomrule
\end{tabular}

\vspace{-2mm}
\end{table*}

%% file: sec/6_conl.tex
\section{Conclusion}
\label{conl}

We presented O$^{2}$LaF, a self-supervised framework
that repurposes a pretrained video DiT for both inverse
and forward dynamics.
By tailoring the two dynamics to their distinct roles
and controlling information access, O$^{2}$LaF learns
transferable latent actions from unlabeled videos.
Across navigation and robotic manipulation, it achieves
state-of-the-art few-shot action probing and enables
zero-shot action transfer across visual contexts and
human--robot embodiments.
These findings establish pretrained video generators
as reusable dynamics priors for action discovery
and controllable world modeling.

\noindent\textbf{Limitations and future work.}
Future work includes improving the interpretability
of the learned latent-action space and extending
video-predictive dynamics to efficient planning
and closed-loop control.
See Appendix~\ref{app:limitations} for further discussion.

%% file: sec/7_supp.tex
\appendixtitle
\suppressfloats[t]

The document provides supplementary information not
elaborated on in our main paper due to space constraints:
implementation
details (Section~\ref{app:details}), visualization of action-query
attention (Section~\ref{app:action_query_attention}) and analysis of training data composition (Section~\ref{sec:dataset_ablation}),
two applications that use the IDM as a motion annotator
and the FDM as a policy verifier (Section~\ref{app:app}), and a
discussion of limitations and future directions
(Section~\ref{app:limitations}).

\section{Implementation Details}
\label{app:details}

\subsection{End-to-End Latent-Action Learning}
\label{app:latent_action_training}


Both the IDM and FDM are initialized from Wan2.2-TI2V-5B~\cite{wan2025wan22},
with frozen video VAE and text encoder.
We adapt both video DiTs using rank-128 LoRA, while
directly training the IDM action queries and Gaussian
readout, and the FDM action embedder.
Each visual transition is represented by a
32-dimensional Gaussian latent action.
The IDM uses frame-causal attention with pairwise action
queries, while the FDM uses frame-causal attention to
denoise future frames conditioned on a clean initial frame.
Training configurations are summarized in
Tab.~\ref{tab:training_details}.

\subsection{Two-Stage Video World Model}
\label{app:two_stage_fdm}
To improve visual generation quality, we freeze the
learned IDM as a latent-action extractor and train a
separate bidirectional Wan2.2-TI2V-5B world model
conditioned on the extracted actions.
The FDM is fully fine-tuned at $832\times480$ resolution.
By default, the FDM is initialized from the pretrained
text/image-to-video model.
For the E2E-init setting, we instead initialize it
from the jointly trained causal FDM, providing an
initialization already aligned with the learned
latent-action space.

\begin{table}[h]
\centering
\caption{
\textbf{Training configurations.}
}
\vspace{-2mm}
\label{tab:training_details}
\small
\setlength{\tabcolsep}{3.5pt}
\renewcommand{\arraystretch}{1.08}
\begin{tabular}{@{}lcc@{}}
\toprule
& \textbf{End-to-end} & \textbf{Two-stage FDM} \\
\midrule
Backbone & Wan2.2-5B $\times 2$ & Wan2.2-5B \\
Trainable weights & LoRA-128 + heads & Full FDM \\
IDM & Jointly trained & Frozen @ 5k \\
Attention & Frame-causal & Bidirectional FDM \\
RGB frames & 41 & 81 \\
Latent actions & 10 & 20 \\
\midrule
Global batch & 256 & 128 \\
GPUs & 16 & 16 \\
Peak GPU memory (GiB) & 65.49 & 55.08 \\
Updates & 5k & 5k \\
Learning rate & $5\times10^{-5}$ & $5\times10^{-5}$ \\
Warmup steps & 50 & 100 \\
\bottomrule
\end{tabular}
\end{table}


\section{More Analysis}

\begin{figure}[t]
  \centering
  \includegraphics[width=\linewidth]{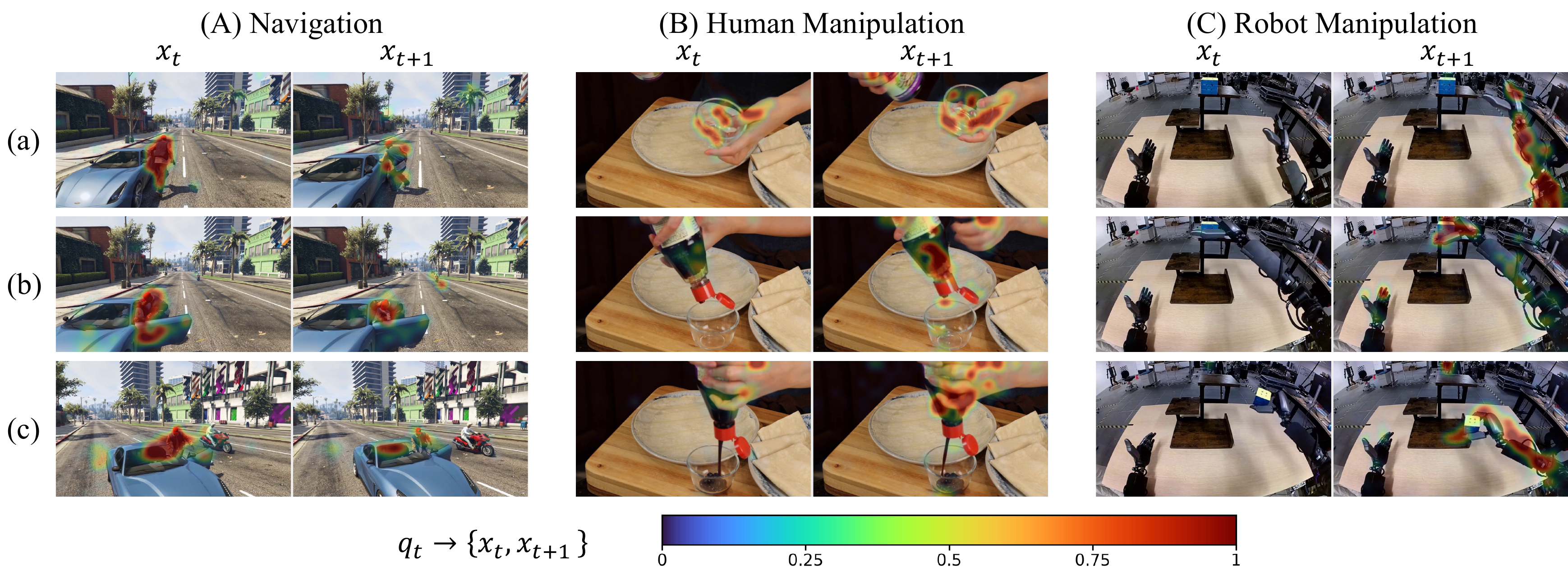}
  \vspace{-5mm}
  \caption{
\textbf{Where does the pairwise action query attend?}
For transition $x_t\!\rightarrow x_{t+1}$, action query $q_t$ jointly attends
to spatial video tokens in both endpoint frames.
}
  \label{supp:fig:viz}
\end{figure}

\begin{table}[t]
\centering
\caption{
\textbf{Effect of latent-action training data.}
All checkpoints are compared at training step 5k.
}
\vspace{-3mm}
\label{tab:dataset_ablation}

\small
\setlength{\tabcolsep}{4.5pt}
\renewcommand{\arraystretch}{1.08}

\begin{tabular}{l|cc|cc}
\toprule
\multirow{2}{*}{\textbf{Training data}}
& \multicolumn{2}{c|}{\textbf{MIND Exact Match} $\uparrow$}
& \multicolumn{2}{c}{\textbf{In-lab EEF} $R^2\uparrow$} \\
\cmidrule(lr){2-3}
\cmidrule(lr){4-5}
& \textsc{1st-P} & \textsc{3rd-P}
& Left arm & Right arm \\
\midrule

Mira only
& \textbf{87.95}
& \textbf{94.37}
& 0.135
& 0.257 \\

EgoDex only
& 76.70
& 86.60
& \underline{0.192}
& \underline{0.363} \\

EgoDexMix
& 76.70
& 91.66
& \textbf{0.227}
& \textbf{0.386} \\

Mira / EgoDexMix (70/30)
& \underline{85.94}
& \underline{93.96}
& 0.163
& 0.341 \\

\bottomrule
\end{tabular}

\vspace{0.4mm}
\end{table}

\subsection{Visualizing Action-Query Attention}
\label{app:action_query_attention}

To understand how the IDM extracts latent actions from visual
transitions, we visualize the attention from learned action queries
to their corresponding source and target frames.
For transition $x_i\!\rightarrow x_{i+1}$, action query $q_i$
jointly attends to the spatial tokens of both frames.
At transformer block $\ell$ and head $h$, we compute
\begin{equation}
\begin{aligned}
[a_{i,\mathrm{src}}^{\ell,h},
 a_{i,\mathrm{tgt}}^{\ell,h}]
=
\operatorname{softmax}\!\left(
\frac{
[q_i^{\ell,h}(K_i^{\ell,h})^\top,\,
 q_i^{\ell,h}(K_{i+1}^{\ell,h})^\top]
}{\sqrt{d_h}}
\right).
\end{aligned}
\end{equation}
The softmax is computed jointly over both frames,
preserving their relative attention mass.
The attention maps have a spatial resolution of $15\times26$
and are bicubically upsampled to the RGB frames.
For each head and video, we normalize attention by the
99th percentile across all transitions and both frames,
followed by clipping to $[0,1]$.
We visualize the resulting maps using the TURBO colormap
with gamma correction ($\gamma=0.8$) and
attention-dependent opacity.
Source and target frames share the same normalization scale,
without frame-wise renormalization or head aggregation.

As shown in Fig.~\ref{supp:fig:viz}, action-query attention
often concentrates on moving agents, manipulated objects,
and hand or end-effector interaction regions.
These attended regions evolve across transitions,
suggesting that the learned queries extract spatially
localized evidence from visual changes.


\subsection{Analyzing Training Data Composition}
\label{sec:dataset_ablation}

Although learned without action annotations, latent-action representations still depend on the motion and embodiment diversity of the training videos.
We compare four configurations: MiraData~\cite{ju2024miradata} (navigation),
EgoDex~\cite{hoque2026egodex} (human manipulation), EgoDexMix (human and robot
manipulation), and joint MiraData/EgoDexMix training.
EgoDexMix augments human demonstrations with real robot
videos from GR-1~\cite{gao2026dreamdojo} and PH2D~\cite{qiu2025humanoid}.
All configurations are evaluated on MIND and In-lab GR-1
following Sec.~\ref{sec:exp:representation}.

Tab.~\ref{tab:dataset_ablation} reveals a trade-off
between domain specialization and cross-domain generalization.
MiraData-only training achieves the strongest navigation
accuracy, whereas EgoDexMix yields the best robot probing.
Notably, compared with human-only EgoDex, EgoDexMix
also improves third-person navigation accuracy.
Joint MiraData/EgoDexMix training largely preserves
navigation performance while improving robot probing
over MiraData alone.
These findings suggest that diversifying training videos
across motion domains and embodiments can broaden
the transferability of learned latent actions.

\section{Applications}
\label{app:app}

\begin{table*}[t]
\centering
\caption{
\textbf{Camera-motion annotation on CameraBench.}
Per-class AP ($\uparrow$) on the official test set.
\emph{Video}/\emph{LA} denote video-generative/latent-action pretraining.
\emph{Frozen-$\mu$}/\emph{Frozen-$h$} use MLP probes on
latent action means/pre-Gaussian features.
\emph{LoRA-$h$}/\emph{Full FT-$h$} additionally adapt the IDM.
Blue and red rows highlight matched supervised comparisons
at 270M and 5B scales, respectively.
$\Delta_{\mathrm{LA}}$ reports the AP gain from action
pretraining over matched baselines.
Our results ensemble three independently trained seeds.
}
\vspace{-3mm}
\label{tab:camera_annotation}

\vspace{0.3em}
\scriptsize
\setlength{\tabcolsep}{2.5pt}
\renewcommand{\arraystretch}{1.12}

\resizebox{\textwidth}{!}{%
\begin{tabular}{lll|*{15}{c}|c|c}
\toprule

\multirow{2}{*}{\textbf{Model}}
& \multirow{2}{*}{\textbf{Pretraining}}
& \multirow{2}{*}{\textbf{Adaptation}}
& \multicolumn{6}{c}{\textbf{Translation (Dolly/Pedestal/Truck)}}
& \multicolumn{2}{c}{\textbf{Zoom}}
& \multicolumn{6}{c}{\textbf{Rotation (Pan/Tilt/Roll)}}
& \multirow{2}{*}{\textbf{Static}}
& \multirow{2}{*}{\textbf{Avg.}}
& \multirow{2}{*}{\textcolor{darkred}{\textbf{$\Delta_{\mathrm{LA}}$}}} \\

\cmidrule(lr){4-9}
\cmidrule(lr){10-11}
\cmidrule(lr){12-17}

& & &
In & Out & Up & Down & Right & Left
& In & Out
& Right & Left & Up & Down & CW & CCW
& & & \\

\midrule

\multicolumn{20}{l}{
\emph{CameraBench reference}
} \\

Qwen2.5-VL-72B
& VLM
& SFT
& 86.8 & 41.3 & 30.5 & 50.6 & 70.7 & 52.6
& 57.1 & 78.5
& 81.9 & 86.6 & 79.1 & 70.4 & 35.0 & 43.8
& 76.6 & 62.8 & -- \\

\midrule

\multicolumn{20}{l}{
\emph{Without latent-action pretraining}
} \\

\rowcolor{blue!8}
Scratch IDM (270M)
& None
& Full FT-$h$
& 45.9 & 12.8 & 9.8 & 15.4 & 32.1 & 21.2
& 14.2 & 12.7
& 35.4 & 32.1 & 29.4 & 22.1 & 15.4 & 14.8
& 66.9 & 25.3 & -- \\

\rowcolor{red!8}
Video DiT IDM (5B)
& Video
& LoRA-$h$
& 93.9 & 72.4 & 54.5 & 73.5 & 81.4 & 77.3
& 68.4 & 76.1
& 91.7 & 90.2 & 81.9 & 71.1 & 88.7 & 90.3
& 87.7 & 79.9 & -- \\

\midrule

\multicolumn{20}{l}{
\emph{With latent-action pretraining}
} \\

O$^{2}$LaF-Scratch (270M)
& LA
& Frozen-$\mu$
& 83.8 & 56.5 & 44.0 & 49.7 & 74.0 & 61.3
& 57.5 & 76.8
& 81.0 & 84.3 & 83.6 & 73.3 & 72.4 & 85.2
& 67.7 & 70.1 & -- \\

\rowcolor{blue!8}
O$^{2}$LaF-Scratch (270M)
& LA
& Full FT-$h$
& 87.5 & 56.4 & 34.6 & 67.3 & 72.5 & 62.8
& 51.8 & 69.4
& 80.7 & 80.5 & 77.4 & 64.1 & 77.6 & 86.6
& 83.5 & 70.2 & \textcolor{darkred}{\textbf{+44.9}} \\

O$^{2}$LaF-DiT (5B)
& Video + LA
& Frozen-$\mu$
& 91.4 & 58.1 & 56.2 & 75.3 & 86.4 & 82.9
& 78.5 & 80.9
& 92.9 & 93.7 & 85.5 & 80.9 & 90.3 & 96.3
& 69.4 & 81.2 & -- \\

O$^{2}$LaF-DiT (5B)
& Video + LA
& Frozen-$h$
& 94.6 & 70.4 & 60.7 & 85.1 & 91.9 & 87.8
& \textbf{83.2} & 84.2
& 95.5 & 93.8 & 87.0 & 80.6 & 92.5 & 95.8
& 81.5 & 85.6 & -- \\

\rowcolor{red!8}
O$^{2}$LaF-DiT (5B)
& Video + LA
& LoRA-$h$
& \textbf{96.4} & \textbf{84.4} & \textbf{67.1}
& \textbf{91.4} & \textbf{95.1} & \textbf{90.0}
& 80.8 & \textbf{86.3}
& \textbf{96.9} & \textbf{95.8} & \textbf{90.1}
& \textbf{83.1} & \textbf{93.5} & \textbf{96.4}
& \textbf{88.6} & \textbf{89.1} & \textcolor{darkred}{\textbf{+9.2}} \\

\bottomrule
\end{tabular}%
}
\vspace{-2mm}
\end{table*}

\subsection{IDM as a Motion Annotator}
\label{sec:motion_annotator}

O$^2$LaF learns latent actions by explaining visual transitions,
without explicit motion annotations.
Can a self-supervised IDM provide a better foundation
for fine-grained motion understanding?
We investigate this on CameraBench~\citep{lin2026towards},
which evaluates 15 camera-motion categories.
We consider two backbone scales: a 270M spatial-temporal
Transformer trained from scratch and fully fine-tuned, and a 5B
video DiT initialized from Wan2.2 and adapted with rank-128 LoRA
(322M trainable parameters).
We compare supervised camera-motion recognition with and without
latent-action pretraining, using $1{,}114$ labeled CameraBench
training clips (1.78 hours).
We additionally probe frozen IDMs with lightweight MLP heads
to assess their motion understanding without backbone adaptation.

As shown in Tab.~\ref{tab:camera_annotation},
latent-action pretraining substantially improves
supervised motion recognition, yielding gains of
44.9 and 9.2 AP for the spatial-temporal Transformer
and pretrained video DiT, respectively.
Notably, the latter improvement comes on top of
strong video generative pretraining, suggesting
that inverse-dynamics learning captures motion
information beyond generic video representations.
Even without backbone adaptation, frozen IDM
features remain highly effective for motion
recognition.
Together, these results establish self-supervised
inverse dynamics as a promising foundation for 
motion annotation, with potential extensions to
dense temporal labeling and feedback for
controllable video generation.




\subsection{FDM as a Policy Verifier}
\label{app:verifier}

Frontier AI models are quickly becoming capable robot agents:
they write robot policies as code~\cite{liang2023codeaspolicies}
and control a two-armed robot from camera images in simulation~\cite{anthropic2026fable}.
Such agents are increasingly built and tested in simulators,
but a simulator only approximates the real scene, and real-robot trials are slow and costly~\cite{zhou2025autoeval},
so a behavior that succeeds in simulation may still fail on the robot.
Our FDM offers a cheap check across this gap:
starting from a real photo, it predicts what a candidate action sequence will do in the real scene,
before the robot executes it.

\begin{figure*}[!t]
  \centering
  \includegraphics[width=0.95\textwidth]{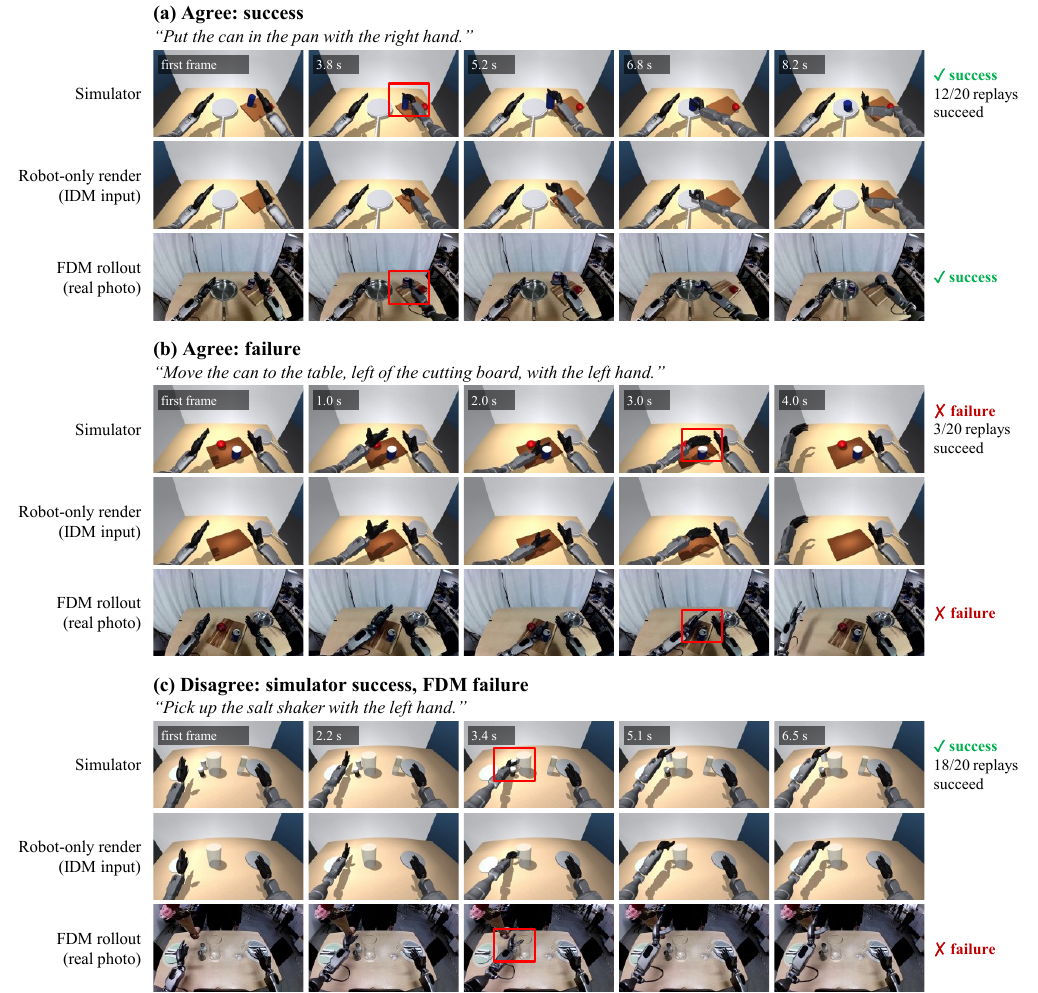}
  \vspace{-2mm}
  \caption{\textbf{The FDM as a policy verifier on a GR-1 humanoid.}
  Each case shows one trajectory in the physics simulator rebuilt from a single photo (top), the same motion with objects removed, from which latent actions are extracted (middle), and the FDM rollout from the real photo (bottom).}
  \label{fig:verifier}
\end{figure*}

For each scene, Opus 5.5 (high) rebuilds a MuJoCo~\cite{todorov2012mujoco} simulation from the robot's head-camera photo, using the GR-1 model from robosuite~\cite{zhu2020robosuite}.
We then script candidate grasps and run them in physics as a reference.
To query the FDM, we render each trajectory with all movable objects removed, so the latent actions the IDM extracts describe the robot's motion but not the simulated outcome, and roll them out from the real photo.
When the two agree (Fig.~\ref{fig:verifier}a,b), the FDM reproduces both a successful placement and a failed grasp, while keeping details the rebuild simplifies, such as the real pan.
When they disagree (Fig.~\ref{fig:verifier}c), the simulator lifts a salt shaker that, in the FDM rollout, stays on the table.
The rebuilt scene suggests a cause: the small tapered glass shakers with metal caps become taller plain cylinders, and the glass vase beside them becomes an opaque cylinder, so a grasp that fits the rebuilt shaker may miss the real one.
Without executing the trajectory we cannot tell which prediction is right, but such a disagreement flags an action worth checking before it reaches the robot.
We see this pre-execution check as a promising direction for future work.

\section{Limitations and Future Work}
\label{app:limitations}

O$^2$LaF learns transferable latent actions from
unlabeled videos, but several challenges remain.

\noindent\textbf{Structure and interpretability of latent actions.}
Although our learned actions transfer across visual
contexts and embodiments, their underlying structure
remains poorly understood. 
Beyond quantitative probing, interpreting individual latent actions still largely relies on inspecting their corresponding reference videos.
Learning more interpretable, compositional, and appearance-invariant action representations remains an important direction for future work.

\noindent\textbf{Toward decision-making world models.}
O$^2$LaF learns latent actions that explain visual transitions and control future video generation.
However, its dynamics are optimized for visual prediction rather than decision-making.
Unlike compact dynamics models that support efficient imagined rollouts and policy optimization~\citep{hafner2025dreamerv3,liao2026supermimic}, video diffusion models incur substantial computational costs and do not directly expose executable actions or task-relevant rewards.
Bridging video-derived action representations with
efficient, decision-relevant dynamics is an important
direction toward planning and closed-loop control.

%% file: ref.bib
@String(CVPR= {IEEE Conf. Comput. Vis. Pattern Recog.})

@String(ICCV= {Int. Conf. Comput. Vis.})

@String(ECCV= {Eur. Conf. Comput. Vis.})

@String(ICLR = {Int. Conf. Learn. Represent.})

@String(AAAI = {AAAI})

@String(CoRL = {Conference on robot learning})

@String(CVPR  = {CVPR})

@String(ICCV  = {ICCV})

@String(ECCV  = {ECCV})

@String(ICLR  = {ICLR})

@String(CoRL = {CoRL})

@String(RSS = {Robotics: Science and Systems})

@article{ha2018worldmodels,
  title={World Models},
  author={Ha, David and Schmidhuber, J{\"u}rgen},
  journal={arXiv preprint arXiv:1803.10122},
  year={2018}
}

@article{hafner2025dreamerv3,
  title={Mastering Diverse Control Tasks through World Models},
  author={Hafner, Danijar and Pasukonis, Jurgis and Ba, Jimmy and Lillicrap, Timothy},
  journal={Nature},
  volume={640},
  number={8059},
  pages={647--653},
  year={2025}
}

@inproceedings{yang2024unisim,
  title={Learning Interactive Real-World Simulators},
  author={Yang, Sherry and Du, Yilun and Ghasemipour, Seyed Kamyar Seyed and Tompson, Jonathan and Kaelbling, Leslie Pack and Schuurmans, Dale and Abbeel, Pieter},
  booktitle={ICLR},
  year={2024}
}

@inproceedings{wu2024ivideogpt,
  title={{iVideoGPT}: Interactive {VideoGPTs} are Scalable World Models},
  author={Wu, Jialong and Yin, Shaofeng and Feng, Ningya and He, Xu and Li, Dong and Hao, Jianye and Long, Mingsheng},
  booktitle={NeurIPS},
  year={2024}
}

@article{nvidia2025cosmos,
  title={{Cosmos} World Foundation Model Platform for Physical {AI}},
  author={{NVIDIA} and Agarwal, Niket and Ali, Arslan and Bala, Maciej and Balaji, Yogesh and Barker, Erik and others},
  journal={arXiv preprint arXiv:2501.03575},
  year={2025}
}

@article{robbyant2026lingbotworld,
  title={Advancing Open-source World Models},
  author={{Robbyant Team}},
  journal={arXiv preprint arXiv:2601.20540},
  year={2026}
}

@misc{parkerholder2025genie3,
  title={{Genie 3}: A New Frontier for World Models},
  author={Parker-Holder, Jack and Fruchter, Shlomi},
  howpublished={Google DeepMind Blog},
  url={https://deepmind.google/discover/blog/genie-3-a-new-frontier-for-world-models/},
  year={2025}
}

@article{rigter2024avid,
  title={Avid: Adapting video diffusion models to world models},
  author={Rigter, Marc and Gupta, Tarun and Hilmkil, Agrin and Ma, Chao},
  journal={arXiv preprint arXiv:2410.12822},
  year={2024}
}

@inproceedings{zitkovich2023rt2,
  title={{RT-2}: Vision-Language-Action Models Transfer Web Knowledge to Robotic Control},
  author={Zitkovich, Brianna and Yu, Tianhe and Xu, Sichun and Xu, Peng and Xiao, Ted and Xia, Fei and Wu, Jialin and Wohlhart, Paul and Welker, Stefan and Wahid, Ayzaan and Vuong, Quan and Vanhoucke, Vincent and Tran, Huong and Soricut, Radu and Singh, Anikait and Singh, Jaspiar and Sermanet, Pierre and Sanketi, Pannag R. and Salazar, Grecia and Ryoo, Michael S. and Reymann, Krista and Rao, Kanishka and Pertsch, Karl and Mordatch, Igor and Michalewski, Henryk and Lu, Yao and Levine, Sergey and Lee, Lisa and Lee, Tsang-Wei Edward and Leal, Isabel and Kuang, Yuheng and Kalashnikov, Dmitry and Julian, Ryan and Joshi, Nikhil J. and Irpan, Alex and Ichter, Brian and Hsu, Jasmine and Herzog, Alexander and Hausman, Karol and Gopalakrishnan, Keerthana and Fu, Chuyuan and Florence, Pete and Finn, Chelsea and Dubey, Kumar Avinava and Driess, Danny and Ding, Tianli and Choromanski, Krzysztof Marcin and Chen, Xi and Chebotar, Yevgen and Carbajal, Justice and Brown, Noah and Brohan, Anthony and Arenas, Montserrat Gonzalez and Han, Kehang},
  booktitle={CoRL},
  year={2023}
}

@article{sima2025sima2,
  title={{SIMA 2}: A Generalist Embodied Agent for Virtual Worlds},
  author={{SIMA Team} and Bolton, Adrian and Lerchner, Alexander and Cordell, Alexandra and Moufarek, Alexandre and Bolt, Andrew and others},
  journal={arXiv preprint arXiv:2512.04797},
  year={2025}
}

@inproceedings{black2025pi0,
  title={$\pi_0$: A Vision-Language-Action Flow Model for General Robot Control},
  author={Black, Kevin and Brown, Noah and Driess, Danny and Esmail, Adnan and Equi, Michael Robert and Finn, Chelsea and Fusai, Niccolo and Groom, Lachy and Hausman, Karol and Ichter, Brian and Jakubczak, Szymon and Jones, Tim and Ke, Liyiming and Levine, Sergey and Li-Bell, Adrian and Mothukuri, Mohith and Nair, Suraj and Pertsch, Karl and Shi, Lucy Xiaoyang and Smith, Laura and Tanner, James and Vuong, Quan and Walling, Anna and Wang, Haohuan and Zhilinsky, Ury},
  booktitle={RSS},
  year={2025}
}

@article{nvidia2025gr00tn1,
  title={{GR00T N1}: An Open Foundation Model for Generalist Humanoid Robots},
  author={{NVIDIA} and Bjorck, Johan and Casta{\~n}eda, Fernando and Cherniadev, Nikita and Da, Xingye and Ding, Runyu and others},
  journal={arXiv preprint arXiv:2503.14734},
  year={2025}
}

@article{gemini2025robotics,
  title={{Gemini Robotics}: Bringing {AI} into the Physical World},
  author={{Gemini Robotics Team} and Abeyruwan, Saminda and Ainslie, Joshua and Alayrac, Jean-Baptiste and Arenas, Montserrat Gonzalez and Armstrong, Travis and others},
  journal={arXiv preprint arXiv:2503.20020},
  year={2025}
}

@article{chen2026hy,
  title={{Hy-Nomas}: Towards Free-form Interactive World Generation},
  author={Chen, Qiaoling and Cheng, Jiaxiang and Deng, Xueqing and Hang, Tiankai and Hou, Ji and Li, Jiaqi and Liu, Yifan and Lu, Tao and Tang, Junshu and Tang, Yu and others},
  journal={arXiv preprint arXiv:2610.36729},
  year={2026}
}

@inproceedings{baker2022vpt,
  title={Video PreTraining ({VPT}): Learning to Act by Watching Unlabeled Online Videos},
  author={Baker, Bowen and Akkaya, Ilge and Zhokhov, Peter and Huizinga, Joost and Tang, Jie and Ecoffet, Adrien and Houghton, Brandon and Sampedro, Raul and Clune, Jeff},
  booktitle={NeurIPS},
  year={2022}
}

@inproceedings{edwards2019ilpo,
  title={Imitating Latent Policies from Observation},
  author={Edwards, Ashley and Sahni, Himanshu and Schroecker, Yannick and Isbell, Charles},
  booktitle={ICML},
  year={2019}
}

@inproceedings{rybkin2019learning,
  title={Learning What You Can Do Before Doing Anything},
  author={Rybkin, Oleh and Pertsch, Karl and Derpanis, Konstantinos G. and Daniilidis, Kostas and Jaegle, Andrew},
  booktitle={ICLR},
  year={2019}
}

@inproceedings{bruce2024genie,
  title={Genie: Generative interactive environments},
  author={Bruce, Jake and Dennis, Michael D and Edwards, Ashley and Parker-Holder, Jack and Shi, Yuge and Hughes, Edward and Lai, Matthew and Mavalankar, Aditi and Steigerwald, Richie and Apps, Chris and others},
  booktitle={ICML},
  year={2024}
}

@inproceedings{gaoadaworld,
  title={{AdaWorld}: Learning Adaptable World Models with Latent Actions},
  author={Gao, Shenyuan and Zhou, Siyuan and Du, Yilun and Zhang, Jun and Gan, Chuang},
  booktitle={ICML},
  year={2025}
}

@inproceedings{
    jiang2026olafworld,
    title={{Olaf-World}: Orienting Latent Actions for Video World Modeling},
    author={Yuxin Jiang and Yuchao Gu and Ivor Tsang and Mike Zheng Shou},
    booktitle={ICML},
    year={2026}
    }

@inproceedings{
wang2026coevolving,
title={{Co-Evolving} Latent Action World Models},
author={Yucen Wang and Fengming Zhang and De-Chuan Zhan and Li Zhao and Kaixin Wang and Jiang Bian},
booktitle={ICML},
year={2026}
}

@inproceedings{
garrido2026learning,
title={Learning Latent Action World Models in the Wild},
author={Quentin Garrido and Tushar Nagarajan and Basile Terver and Nicolas Ballas and Yann LeCun and Michael Rabbat},
booktitle={ICML},
year={2026}
}

@inproceedings{schmidt2024lapo,
  title={Learning to Act without Actions},
  author={Schmidt, Dominik and Jiang, Minqi},
  booktitle={ICLR},
  year={2024}
}

@inproceedings{ye2024lapa,
  title={Latent Action Pretraining from Videos},
  author={Ye, Seonghyeon and Jang, Joel and Jeon, Byeongguk and Joo, Sejune and Yang, Jianwei and Peng, Baolin and Mandlekar, Ajay and Tan, Reuben and Chao, Yu-Wei and Lin, Bill Yuchen and Liden, Lars and Lee, Kimin and Gao, Jianfeng and Zettlemoyer, Luke and Fox, Dieter and Seo, Minjoon},
  booktitle={ICLR},
  year={2025}
}

@inproceedings{chen2024moto,
  title={{Moto}: Latent Motion Token as the Bridging Language for Learning Robot Manipulation from Videos},
  author={Chen, Yi and Ge, Yuying and Tang, Weiliang and Li, Yizhuo and Ge, Yixiao and Ding, Mingyu and Shan, Ying and Liu, Xihui},
  booktitle={ICCV},
  year={2025}
}

@inproceedings{bu2025univla,
  title={{UniVLA}: Learning to Act Anywhere with Task-centric Latent Actions},
  author={Bu, Qingwen and Yang, Yanting and Cai, Jisong and Gao, Shenyuan and Ren, Guanghui and Yao, Maoqing and Luo, Ping and Li, Hongyang},
  booktitle={RSS},
  year={2025}
}

@inproceedings{kim2025uniskill,
  title={{UniSkill}: Imitating Human Videos via Cross-Embodiment Skill Representations},
  author={Kim, Hanjung and Kang, Jaehyun and Kang, Hyolim and Cho, Meedeum and Kim, Seon Joo and Lee, Youngwoon},
  booktitle={CoRL},
  year={2025}
}

@inproceedings{yang2026como,
  title={{CoMo}: Learning Continuous Latent Motion from Internet Videos for Scalable Robot Learning},
  author={Yang, Jiange and Shi, Yansong and Zhu, Haoyi and Liu, Mingyu and Ma, Kaijing and Wang, Yating and Wu, Gangshan and He, Tong and Wang, Limin},
  booktitle={CVPR},
  year={2026}
}

@inproceedings{gao2026dreamdojo,
  title={{DreamDojo}: A Generalist Robot World Model from Large-Scale Human Videos},
  author={Gao, Shenyuan and Liang, William and Zheng, Kaiyuan and Malik, Ayaan and Ye, Seonghyeon and Yu, Sihyun and Tseng, Wei-Cheng and Dong, Yuzhu and Mo, Kaichun and Lin, Chen-Hsuan and Ma, Qianli and Nah, Seungjun and Magne, Loic and Xiang, Jiannan and Xie, Yuqi and Zheng, Ruijie and Niu, Dantong and Tan, You Liang and Zentner, K.R. and Kurian, George and Indupuru, Suneel and Jannaty, Pooya and Gu, Jinwei and Zhang, Jun and Malik, Jitendra and Abbeel, Pieter and Liu, Ming-Yu and Zhu, Yuke and Jang, Joel and Fan, Linxi},
  booktitle={ICML},
  year={2026}
}

@inproceedings{ren2026videoworld2,
  title={{VideoWorld 2}: Learning Transferable Knowledge from Real-world Videos},
  author={Ren, Zhongwei and Wei, Yunchao and Yu, Xiao and Luo, Guixun and Zhao, Yao and Kang, Bingyi and Feng, Jiashi and Jin, Xiaojie},
  booktitle={CVPR},
  year={2026}
}

@inproceedings{bauer2026lad,
  title={Latent Action Diffusion for Cross-Embodiment Manipulation},
  author={Bauer, Erik and Nava, Elvis and Katzschmann, Robert K.},
  booktitle={ICRA},
  year={2026}
}

@inproceedings{chen2026villax,
  title={{villa-X}: Enhancing Latent Action Modeling in Vision-Language-Action Models},
  author={Chen, Xiaoyu and Wei, Hangxing and Zhang, Pushi and Zhang, Chuheng and Wang, Kaixin and Guo, Yanjiang and Yang, Rushuai and Wang, Yucen and Xiao, Xinquan and Zhao, Li and Chen, Jianyu and Bian, Jiang},
  booktitle={ICLR},
  year={2026}
}

@inproceedings{lee2026mvplam,
  title={{MVP-LAM}: Learning Action-Centric Latent Action via Cross-Viewpoint Reconstruction},
  author={Lee, Jung Min and Lee, Dohyeok and Ju, Seokhun and Cho, Taehyun and Koo, Jin Woo and Zhao, Li and Hong, Sangwoo and Lee, Jungwoo},
  booktitle={ICML},
  year={2026}
}

@inproceedings{huang2026lacwm,
  title={Cross-Embodiment Robot Foundation World Models with Latent Actions},
  author={Huang, Huang and Yenamandra, Sriram and Majumdar, Arjun and Aljalbout, Elie and Nagarajan, Tushar and Yang, Tsung-Yen and Rai, Akshara and Rabbat, Michael and Fei-Fei, Li and Wu, Jiajun and Wu, Tingfan and Meier, Franziska},
  booktitle={ICML},
  year={2026}
}

@article{yao2026unilap,
  title={{UniLAP}: A Unified Latent Action Policy for Cross-Embodiment Manipulation},
  author={Yao, Lei and Feng, Qiao and Tong, Mutian and Yun, Qihao and Chau, Lap-Pui and Liu, Lingjie},
  journal={arXiv preprint arXiv:2610.16969},
  year={2026}
}

@article{wang2026repwam,
  title={{RepWAM}: World Action Modeling with Representation Visual-Action Tokenizers},
  author={Wang, Junke and Zhang, Qihang and Yang, Shuai and Luo, Yiming and Shen, Yujun and Wu, Zuxuan and Jiang, Yu-Gang and Xu, Yinghao},
  journal={arXiv preprint arXiv:2606.13674},
  year={2026}
}

@article{zhang2026clap,
  title={{CLAP}: Contrastive Latent Action Pretraining for Learning Vision-Language-Action Models from Human Videos},
  author={Zhang, Chubin and Wang, Jianan and Gao, Zifeng and Su, Yue and Dai, Tianru and Zhou, Cai and Lu, Jiwen and Tang, Yansong},
  journal={arXiv preprint arXiv:2601.04061},
  year={2026}
}

@inproceedings{bu2026laof,
  title={{LAOF}: Robust Latent Action Learning with Optical Flow Constraints},
  author={Bu, Xizhou and Lyu, Jiexi and Sun, Fulei and Yang, Ruichen and Ma, Zhiqiang and Li, Wei},
  booktitle={CVPR},
  year={2026}
}

@article{liu2026scar,
  title={{SCAR}: Self-Supervised Continuous Action Representation Learning},
  author={Liu, Hongjia and Feng, Fan and Fu, Minghao and Wang, Xinyue and Lu, Haofei and Huang, Biwei},
  journal={arXiv preprint arXiv:2605.16412},
  year={2026}
}

@article{bu2026whatmatters,
  title={What Matters for Latent Actions in Robot Learning},
  author={Bu, Xizhou and Hu, Qingda and Zhou, Lei and Zhang, Lingfeng and Tang, Yingbo and Liu, Zihao and Tao, Xinyi and Ma, Zhiqiang and Huang, Qingqiu and Tang, Chufeng and Wang, Hongbo and Zhang, Jing and Ma, Jiayi and Ye, Hangjun and Li, Wei and Hao, Xiaoshuai},
  journal={arXiv preprint arXiv:2608.19613},
  year={2026}
}

@inproceedings{klepach2026objectcentric,
  title={Object-Centric Latent Action Learning},
  author={Klepach, Albina and Nikulin, Alexander and Zisman, Ilya and Tarasov, Denis and Derevyagin, Alexander and Polubarov, Andrei and Lyubaykin, Nikita and Kiselev, Igor and Kurenkov, Vladislav},
  booktitle={AAAI},
  year={2026}
}

@article{dai2026conla,
  title={{ConLA}: Contrastive Latent Action Learning from Human Videos for Robotic Manipulation},
  author={Dai, Weisheng and Lan, Kai and Zhou, Jianyi and Zhao, Bo and Su, Xiu and Tong, Junwen and Guan, Weili and Yang, Shuo},
  journal={arXiv preprint arXiv:2602.00557},
  year={2026}
}

@article{yang2026lawm3d,
  title={{LAWM-3D}: Learning {3D}-Aware Latent Actions from Human Videos for Generalizable Robot World Models},
  author={Yang, Jiarui and Zhange, Jiale and Li, Jiawei and Guo, Hang and Huang, Wen and Wang, Jinpeng and Liu, Peidong and Xia, Shu-Tao},
  journal={arXiv preprint arXiv:2608.05706},
  year={2026}
}

@inproceedings{bi2026motus,
  title={{Motus}: A Unified Latent Action World Model},
  author={Bi, Hongzhe and Tan, Hengkai and Xie, Shenghao and Wang, Zeyuan and Huang, Shuhe and Liu, Haitian and Zhao, Ruowen and Feng, Yao and Xiang, Chendong and Rong, Yinze and Zhao, Hongyan and Liu, Hanyu and Su, Zhizhong and Ma, Lei and Su, Hang and Zhu, Jun},
  booktitle={CVPR},
  year={2026}
}

@inproceedings{nie2026lary,
  title={{LARY}: A Latent Action Representation Yielding Benchmark for Generalizable Vision-to-Action Alignment},
  author={Nie, Dujun and Chen, Fengjiao and Lv, Qi and Kuang, Jun and Li, Xiaoyu and Cao, Xuezhi and Cai, Xunliang},
  booktitle={ECCV},
  year={2026}
}

@inproceedings{jeong2026viewinvariant,
  title={Learning to Act Robustly with View-Invariant Latent Actions},
  author={Jeong, Youngjoon and Chun, Junha and Kim, Taesup},
  booktitle={CVPR},
  year={2026}
}

@article{assran2025vjepa2,
  title={{V-JEPA 2}: Self-Supervised Video Models Enable Understanding, Prediction and Planning},
  author={Assran, Mido and Bardes, Adrien and Fan, David and Garrido, Quentin and Howes, Russell and Komeili, Mojtaba and Muckley, Matthew and Rizvi, Ammar and others},
  journal={arXiv preprint arXiv:2506.09985},
  year={2025}
}

@inproceedings{wang2026genception,
  title={Video Generation Models are General-Purpose Vision Learners},
  author={Wang, Letian and Zhang, Chuhan and Kabra, Rishabh and Uijlings, Jasper and Waslander, Steven and Zisserman, Andrew and Carreira, Joao and He, Kaiming and Andriluka, Misha and Bazavan, Eduard Gabriel and Zanfir, Andrei and Sminchisescu, Cristian},
  booktitle={ECCV},
  year={2026}
}

@article{nam2026trackcraft3r,
  title={{TrackCraft3R}: Repurposing Video Diffusion Transformers for Dense 3D Tracking},
  author={Nam, Jisu and Koo, Jahyeok and Son, Soowon and Jung, Jaewoo and An, Honggyu and Hur, Junhwa and Kim, Seungryong},
  journal={arXiv preprint arXiv:2605.12587},
  year={2026}
}

@inproceedings{difftrack,
  title={Emergent Temporal Correspondences from Video Diffusion Transformers},
  author={Nam, Jisu and Son, Soowon and Chung, Dahyun and Kim, Jiyoung and Jin, Siyoon and Hur, Junhwa and Kim, Seungryong},
  booktitle={NeurIPS},
  year={2025}
}

@inproceedings{ditflow,
  title={Video Motion Transfer with Diffusion Transformers},
  author={Pondaven, Alexander and Siarohin, Aliaksandr and Tulyakov, Sergey and Torr, Philip and Pizzati, Fabio},
  booktitle={CVPR},
  year={2025}
}

@article{ditracker,
  title={Probing and Leveraging Video Diffusion Transformer Features for Robust Point Tracking},
  author={Son, Soowon and An, Honggyu and Nam, Jisu and Ko, Hyunah and Kim, Chaehyun and Chung, Dahyun and Jin, Siyoon and Yi, Jung and Hur, Junhwa and Kim, Seungryong},
  journal={arXiv preprint arXiv:2512.20606},
  year={2025}
}

@inproceedings{tinyhistory,
  title={{TinyHistory}: Lightweight Video History Embeddings via Two-Stage Context Learning},
  author={Zhang, Lvmin and Cai, Shengqu and Li, Muyang and Zeng, Chong and Lu, Beijia and Rao, Anyi and Han, Song and Wetzstein, Gordon and Agrawala, Maneesh},
  booktitle={ECCV},
  year={2026}
}

@inproceedings{vpp,
  title={Video Prediction Policy: A Generalist Robot Policy with Predictive Visual Representations},
  author={Hu, Yucheng and Guo, Yanjiang and Wang, Pengchao and Chen, Xiaoyu and Wang, Yen-Jen and Zhang, Jianke and Sreenath, Koushil and Lu, Chaochao and Chen, Jianyu},
  booktitle={ICML},
  year={2025}
}

@article{vpp2,
  title={Video Prediction Policy 2: Predict Better, Act Better},
  author={Guo, Yanjiang and Yan, Haodong and Zhong, Zhide and Zhang, Zhongru and Yang, Qingyuan and Lu, Qingzhou and Chen, Xiaoyu and Wang, Yen-Jen and Deng, Shuying and Yang, Chenghan and Yuan, Puzhen and Liu, Chenxin and Ban, Tun and Zhu, Xiang and Liu, Yichen and Feng, Kun and Li, Haoang and Chen, Jianyu},
  journal={arXiv preprint arXiv:2610.10270},
  year={2026}
}

@inproceedings{cosmospolicy,
  title={{Cosmos Policy}: Fine-Tuning Video Models for Visuomotor Control and Planning},
  author={Kim, Moo Jin and Gao, Yihuai and Lin, Tsung-Yi and Lin, Yen-Chen and Ge, Yunhao and Lam, Grace and Liang, Percy and Song, Shuran and Liu, Ming-Yu and Finn, Chelsea and Gu, Jinwei},
  booktitle={ICLR},
  year={2026}
}

@article{dreamzero,
  title={World Action Models are Zero-shot Policies},
  author={Ye, Seonghyeon and Ge, Yunhao and Zheng, Kaiyuan and Gao, Shenyuan and Yu, Sihyun and Kurian, George and Indupuru, Suneel and Tan, You Liang and Zhu, Chuning and Xiang, Jiannan and Malik, Ayaan and Lee, Kyungmin and Liang, William and Ranawaka, Nadun and Gu, Jiasheng and Xu, Yinzhen and Wang, Guanzhi and Hu, Fengyuan and Narayan, Avnish and Bjorck, Johan and Wang, Jing and Kim, Gwanghyun and Niu, Dantong and Zheng, Ruijie and Xie, Yuqi and Wu, Jimmy and Wang, Qi and Julian, Ryan and Xu, Danfei and Du, Yilun and Chebotar, Yevgen and Reed, Scott and Kautz, Jan and Zhu, Yuke and Fan, Linxi and Jang, Joel},
  journal={arXiv preprint arXiv:2602.15922},
  year={2026}
}

@article{fastwam,
  title={{Fast-WAM}: Do World Action Models Need Test-time Future Imagination?},
  author={Yuan, Tianyuan and Dong, Zibin and Liu, Yicheng and Zhao, Hang},
  journal={arXiv preprint arXiv:2603.16666},
  year={2026}
}

@article{ma2026dit4dit,
  title={{DiT4DiT}: Jointly Modeling Video Dynamics and Actions for Generalizable Robot Control},
  author={Ma, Teli and Zheng, Jia and Wang, Zifan and Jiang, Chunli and Cui, Andy and Liang, Junwei and Yang, Shuo},
  journal={arXiv preprint arXiv:2603.10448},
  year={2026}
}

@inproceedings{peebles2023dit,
  title={Scalable Diffusion Models with Transformers},
  author={Peebles, William and Xie, Saining},
  booktitle={ICCV},
  year={2023}
}

@inproceedings{lipman2023flowmatching,
  title={Flow Matching for Generative Modeling},
  author={Lipman, Yaron and Chen, Ricky T. Q. and Ben-Hamu, Heli and Nickel, Maximilian and Le, Matthew},
  booktitle={ICLR},
  year={2023}
}

@inproceedings{hu2022lora,
  title={{LoRA}: Low-Rank Adaptation of Large Language Models},
  author={Hu, Edward J. and Shen, Yelong and Wallis, Phillip and Allen-Zhu, Zeyuan and Li, Yuanzhi and Wang, Shean and Wang, Lu and Chen, Weizhu},
  booktitle={ICLR},
  year={2022}
}

@article{esser2024scaling,
  title   = {Scaling Rectified Flow Transformers for High-Resolution Image Synthesis},
  author  = {Esser, Patrick and Kulal, Sumith and Blattmann, Andreas and Entezari, Rahim
             and M{\"u}ller, Jonas and Saini, Harry and Levi, Yam and Lorenz, Dominik
             and Sauer, Axel and Boesel, Frederic and Podell, Dustin and Dockhorn, Tim
             and English, Zion and Lacey, Kyle and Goodwin, Alex and Marek, Yannik
             and Rombach, Robin},
  journal = {arXiv preprint arXiv:2403.03206},
  year    = {2024}
}

@misc{diffsynthstudio,
  author       = {{ModelScope Team}},
  title        = {{DiffSynth-Studio}: An Open-Source Diffusion Model Engine},
  year         = {2024},
  url          = {https://github.com/modelscope/DiffSynth-Studio},
  note         = {GitHub repository}
}

@inproceedings{yang2025cogvideox,
  title={{CogVideoX}: Text-to-Video Diffusion Models with An Expert Transformer},
  author={Yang, Zhuoyi and Teng, Jiayan and Zheng, Wendi and Ding, Ming and Huang, Shiyu and Xu, Jiazheng and Yang, Yuanming and Hong, Wenyi and Zhang, Xiaohan and Feng, Guanyu and Yin, Da and Zhang, Yuxuan and Wang, Weihan and Cheng, Yean and Xu, Bin and Gu, Xiaotao and Dong, Yuxiao and Tang, Jie},
  booktitle={ICLR},
  year={2025}
}

@article{polyak2024moviegen,
  title={{Movie Gen}: A Cast of Media Foundation Models},
  author={Polyak, Adam and Zohar, Amit and Brown, Andrew and Tjandra, Andros and Sinha, Animesh and others},
  journal={arXiv preprint arXiv:2410.13720},
  year={2024}
}

@article{kong2024hunyuanvideo,
  title={{HunyuanVideo}: A Systematic Framework For Large Video Generative Models},
  author={Kong, Weijie and Tian, Qi and Zhang, Zijian and Min, Rox and Dai, Zuozhuo and others},
  journal={arXiv preprint arXiv:2412.03603},
  year={2024}
}

@article{wan2025wan,
  title={{Wan}: Open and Advanced Large-Scale Video Generative Models},
  author={{Team Wan} and Wang, Ang and Ai, Baole and Wen, Bin and Mao, Chaojie and others},
  journal={arXiv preprint arXiv:2503.20314},
  year={2025}
}

@misc{wan2025wan22,
  title={{Wan2.2}: Open and Advanced Large-Scale Video Generative Models},
  author={{Team Wan}},
  howpublished={\url{https://github.com/Wan-Video/Wan2.2}},
  year={2025}
}

@article{seedance2026seedance2,
  title={{Seedance 2.0}: Advancing Video Generation for World Complexity},
  author={{Team Seedance} and Chen, De and Chen, Liyang and Chen, Xin and Chen, Ying and others},
  journal={arXiv preprint arXiv:2604.14148},
  year={2026}
}

@misc{brooks2024sora,
  title={Video Generation Models as World Simulators},
  author={Brooks, Tim and Peebles, Bill and Holmes, Connor and DePue, Will and Guo, Yufei and Jing, Li and Schnurr, David and Taylor, Joe and Luhman, Troy and Luhman, Eric and Ng, Clarence and Wang, Ricky and Ramesh, Aditya},
  howpublished={OpenAI technical report},
  url={https://openai.com/index/video-generation-models-as-world-simulators/},
  year={2024}
}

@misc{minimax2026h3,
  title={{MiniMax H3}: An Open Model Breaking the Boundaries Between Tasks and Modalities},
  author={{MiniMax}},
  howpublished={MiniMax Blog},
  url={https://www.minimax.io/blog/minimax-h3},
  year={2026}
}

@inproceedings{kingma2014vae,
  title={Auto-Encoding Variational Bayes},
  author={Kingma, Diederik P. and Welling, Max},
  booktitle={ICLR},
  year={2014}
}

@inproceedings{higgins2017betavae,
  title={{$\beta$-VAE}: Learning Basic Visual Concepts with a Constrained Variational Framework},
  author={Higgins, Irina and Matthey, Lo{\"i}c and Pal, Arka and Burgess, Christopher and Glorot, Xavier and Botvinick, Matthew and Mohamed, Shakir and Lerchner, Alexander},
  booktitle={ICLR},
  year={2017}
}

@article{ye2026mind,
  title={{MIND}: Benchmarking Memory Consistency and Action Control in World Models},
  author={Ye, Yixuan and Lu, Xuanyu and Jiang, Yuxin and Gu, Yuchao and Zhao, Rui and Liang, Qiwei and Pan, Jiachun and Zhang, Fengda and Wu, Weijia and Wang, Alex Jinpeng},
  journal={arXiv preprint arXiv:2602.08025},
  year={2026}
}

@inproceedings{ju2024miradata,
  title={{MiraData}: A Large-Scale Video Dataset with Long Durations and Structured Captions},
  author={Ju, Xuan and Gao, Yiming and Zhang, Zhaoyang and Yuan, Ziyang and Wang, Xintao and Zeng, Ailing and Xiong, Yu and Xu, Qiang and Shan, Ying},
  booktitle={NeurIPS Datasets and Benchmarks Track},
  year={2024}
}

@inproceedings{hoque2026egodex,
  title={{EgoDex}: Learning Dexterous Manipulation from Large-Scale Egocentric Video},
  author={Hoque, Ryan and Huang, Peide and Yoon, David and Sivapurapu, Mouli and Zhang, Jian},
  booktitle={ICLR},
  year={2026}
}

@article{qiu2025humanoid,
  title={Humanoid policy\~{} human policy},
  author={Qiu, Ri-Zhao and Yang, Shiqi and Cheng, Xuxin and Chawla, Chaitanya and Li, Jialong and He, Tairan and Yan, Ge and Yoon, David J and Hoque, Ryan and Paulsen, Lars and others},
  journal={arXiv preprint arXiv:2503.13441},
  year={2025}
}

@article{li2026videochat3,
  title={{VideoChat3}: Fully Open Video {MLLM} for Efficient and Generalist Video Understanding},
  author={Li, Xinhao and Zhu, Yuhan and Zeng, Xiangyu and Dong, Yuhao and Wu, Haoning and Zhang, Zhiqiu and Yang, Yuandong and Ma, Changlian and Zhang, Qingyu and Shi, Yansong and Chen, Xinyu and Chen, Haoran and Huang, Zizheng and Zhang, Jun and Ouyang, Kun and Sui, Lin and Yan, Ziang and Xu, Yicheng and Wang, Chenting and He, Yinan and Zhang, Hongjie and Wang, Yi and Qiao, Yu and Wang, Yali and Liu, Ziwei and Chen, Kai and Wang, Limin},
  journal={arXiv preprint arXiv:2607.14935},
  year={2026}
}

@article{zhang2022light,
  title={Light-weight Probing of Unsupervised Representations for Reinforcement Learning},
  author={Zhang, Wancong and GX-Chen, Anthony and Sobal, Vlad and LeCun, Yann and Carion, Nicolas},
  journal={arXiv preprint arXiv:2208.12345},
  year={2022}
}

@inproceedings{huang2024vbench,
  title={{VBench}: Comprehensive Benchmark Suite for Video Generative Models},
  author={Huang, Ziqi and He, Yinan and Yu, Jiashuo and Zhang, Fan and Si, Chenyang and Jiang, Yuming and Zhang, Yuanhan and Wu, Tianxing and Jin, Qingyang and Chanpaisit, Nattapol and Wang, Yaohui and Chen, Xinyuan and Wang, Limin and Lin, Dahua and Qiao, Yu and Liu, Ziwei},
  booktitle={CVPR},
  year={2024}
}

@article{huang2025vipe,
  title={{ViPE}: Video Pose Engine for {3D} Geometric Perception},
  author={Huang, Jiahui and Zhou, Qunjie and Rabeti, Hesam and Korovko, Aleksandr and Ling, Huan and Ren, Xuanchi and Shen, Tianchang and Gao, Jun and Slepichev, Dmitry and Lin, Chen-Hsuan and Ren, Jiawei and Xie, Kevin and Biswas, Joydeep and Leal-Taix{\'e}, Laura and Fidler, Sanja},
  journal={arXiv preprint arXiv:2508.10934},
  year={2025}
}

@article{ho2022cfg,
  title={Classifier-Free Diffusion Guidance},
  author={Ho, Jonathan and Salimans, Tim},
  journal={arXiv preprint arXiv:2207.12598},
  year={2022}
}

@article{tang2023emergent,
  title={Emergent correspondence from image diffusion},
  author={Tang, Luming and Jia, Menglin and Wang, Qianqian and Phoo, Cheng Perng and Hariharan, Bharath},
  journal={NeurIPS},
  year={2023}
}

@article{chen2024diffusion,
  title={Diffusion Forcing: Next-token prediction meets full-sequence diffusion},
  author={Chen, Boyuan and Mart{\'\i} Mons{\'o}, Diego and Du, Yilun and Simchowitz, Max and Tedrake, Russ and Sitzmann, Vincent},
  journal={NeurIPS},
  year={2024}
}

@inproceedings{yang2026stablevelocity,
  title     = {Stable Velocity: A Variance Perspective on Flow Matching},
  author    = {Yang, Donglin and Zhang, Yongxing and Yu, Xin
               and Hou, Liang and Tao, Xin and Wan, Pengfei
               and Qi, Xiaojuan and Liao, Renjie},
  booktitle = {ICML},
  year      = {2026}
}

@article{lin2026towards,
  title={Towards understanding camera motions in any video},
  author={Lin, Zhiqiu and Cen, Siyuan and Jiang, Daniel and Karhade, Jay and Wang, Hewei and Mitra, Chancharik and Ling, Yu Tong Tiffany and Huang, Yuhan and Zawar, Rushikesh and Bai, Xue and others},
  journal={NeurIPS},
  year={2026}
}

@article{liao2026supermimic,
  title={{SuperMimic:} Scaling humanoid whole-body control with one differentiable world model},
  author={Liao, Qiayuan and Shao, Yiyang and Long, Junfeng and Zakka, Kevin and Horowitz, Roberto and Abbeel, Pieter and Sreenath, Koushil},
  journal={arXiv preprint arXiv:2610.28480},
  year={2026}
}

@article{liang2023codeaspolicies,
  title   = {Code as Policies: Language Model Programs for Embodied Control},
  author  = {Liang, Jacky and Huang, Wenlong and Xia, Fei and Xu, Peng and Hausman, Karol and Ichter, Brian and Florence, Pete and Zeng, Andy},
  journal = {arXiv preprint arXiv:2209.07753},
  year    = {2022}
}

@misc{anthropic2026fable,
  title        = {{Claude Fable 5.2 and Claude Mythos 5.2}},
  author       = {{Anthropic}},
  year         = {2026},
  howpublished = {\url{https://www.anthropic.com/claude-fable-and-mythos-5-2}},
}

@inproceedings{zhou2025autoeval,
  title     = {{AutoEval}: Autonomous Evaluation of Generalist Robot Manipulation Policies in the Real World},
  author    = {Zhou, Zhiyuan and Atreya, Pranav and Tan, You Liang and Pertsch, Karl and Levine, Sergey},
  booktitle = {CoRL},
  year      = {2025}
}

@inproceedings{todorov2012mujoco,
  title     = {{MuJoCo}: A physics engine for model-based control},
  author    = {Todorov, Emanuel and Erez, Tom and Tassa, Yuval},
  booktitle = {IROS},
  pages     = {5026--5033},
  year      = {2012}
}

@article{zhu2020robosuite,
  title   = {robosuite: A Modular Simulation Framework and Benchmark for Robot Learning},
  author  = {Zhu, Yuke and Wong, Josiah and Mandlekar, Ajay and Mart{\'i}n-Mart{\'i}n, Roberto and Joshi, Abhishek and Nasiriany, Soroush and Zhu, Yifeng},
  journal = {arXiv preprint arXiv:2009.12293},
  year    = {2020}
}
